\documentclass{article} %
\usepackage{arxiv_conference,times}

\usepackage{amsmath,amsfonts,bm}

\def\eqref#1{equation~\ref{#1}}

\def\1{\bm{1}}

\DeclareMathAlphabet{\mathsfit}{\encodingdefault}{\sfdefault}{m}{sl}
\SetMathAlphabet{\mathsfit}{bold}{\encodingdefault}{\sfdefault}{bx}{n}

\usepackage{xcolor}
\usepackage{soul}
\usepackage{graphicx}
\usepackage{wrapfig}
\usepackage[caption=false]{subfig}
\usepackage{float}
\usepackage{booktabs}
\usepackage{tabularx}
\usepackage{multirow}
\usepackage{colortbl}
\usepackage{pifont}
\usepackage{capt-of}
\usepackage{hyperref}
\usepackage{url}
\usepackage{listings}

\lstdefinestyle{promptstyle}{
  basicstyle=\ttfamily\scriptsize,
  breaklines=true,
  breakindent=0pt,
  breakatwhitespace=true,
  columns=fullflexible,
  keepspaces=true,
  showstringspaces=false,
  xleftmargin=0.5em,
  frame=none,
  literate={\`}{{\textasciigrave}}1
}

\title{CoDimRecon: Agentic Reconstruction of Sim-Ready 3D Scenes with Deformable Curves, Surfaces, and Volumes}

\author{Shuzhao Xie$^{1,2*}$, Lelin Wang$^{2*}$, Guying Lin$^{2}$, Zhi Wang$^{1\dagger}$, Minchen Li$^{2,3\dagger}$ \\
$^{1}$SIGS, Tsinghua University \quad $^{2}$Carnegie Mellon University \quad $^{3}$Genesis AI \\
{\hypersetup{pdfborder={0 0 0}}\textcolor{linkpink}{\url{https://shuzhaoxie.github.io/CoDimRecon}}}
}

\definecolor{todobluebg}{RGB}{218,238,255}
\definecolor{todobluefg}{RGB}{0,77,153}
\definecolor{darkgreen}{rgb}{0,0.5,0}
\definecolor{mainblue}{HTML}{00B0F0}
\definecolor{linkpink}{RGB}{231,70,151}
\colorlet{gbest}{mainblue!45}
\colorlet{gsecond}{mainblue!15}

\newcommand{\cmark}{\textcolor{darkgreen}{\ding{51}}}
\newcommand{\xmark}{\textcolor{red}{\ding{55}}}

\iclrfinalcopy
\begin{document}

\maketitle
{\renewcommand{\thefootnote}{\fnsymbol{footnote}}\footnotetext[1]{Equal contribution. \quad $^{\dagger}$Corresponding authors.}}
\vspace{-0.1in}

\begin{figure}[H]
    \centering
    \includegraphics[width=\linewidth]{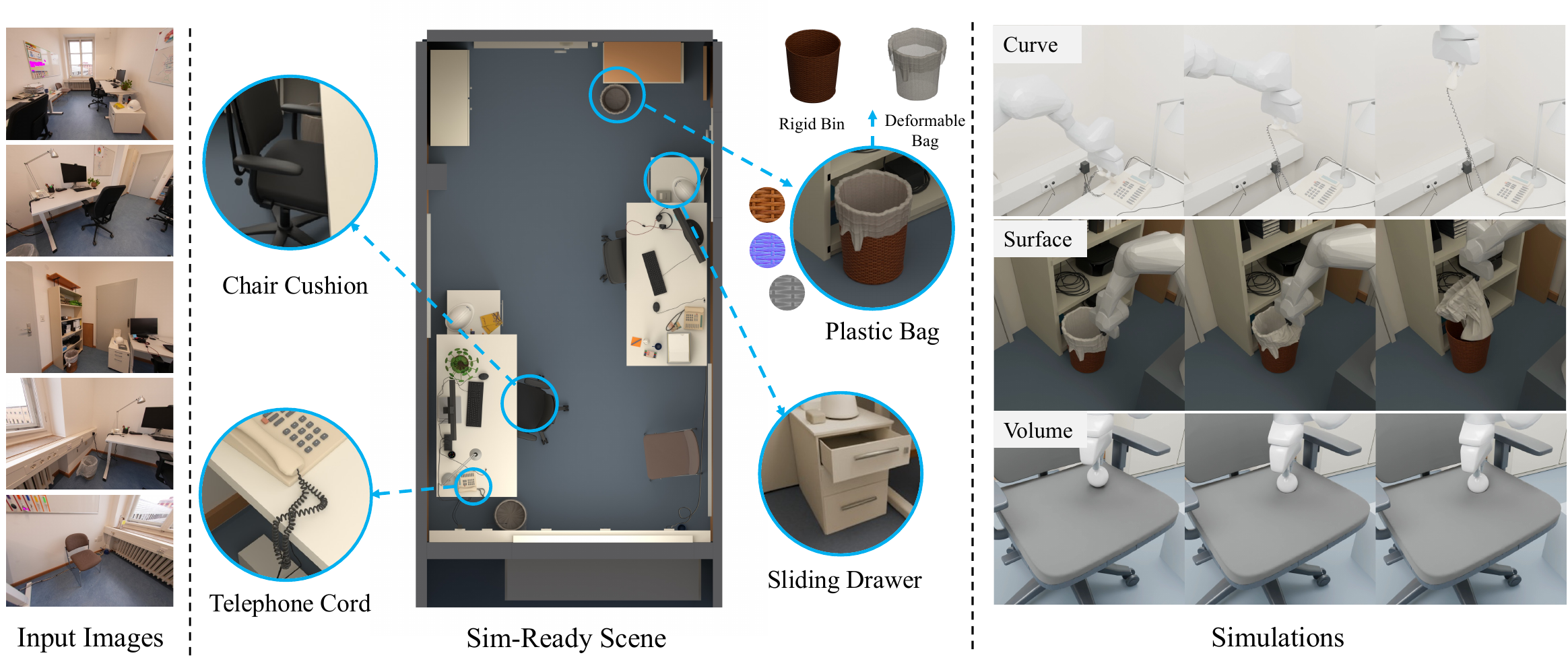}
    \caption{\textbf{CoDimRecon reconstructs a simulation-ready scene from multi-view RGB images.} Left: input views of an office. Center: top-down reconstruction with an articulated drawer and deformable examples---a coiled telephone cord (curve), plastic bag (surface), and chair cushion (volume). The bin and the bag are separate objects, rigid and deformable, respectively. Right: robot interactions with the reconstructed cord, plastic bag, and cushion.}
    \label{fig:teaser}
\end{figure}

\begin{abstract}
Reconstructing simulation-ready 3D scenes from real-world observations enables robotics, gaming, and immersive applications, yet existing methods largely assume rigid objects. This leaves an important gap for deformables, whose simulation-ready geometry depends on dimensionality---curves, surfaces, or volumes---and whose behavior may require models beyond elasticity. We present CoDimRecon, an agentic framework that reconstructs editable scenes containing rigid, articulated, and deformable objects from multi-view RGB observations. Scene-level geometric priors ground scale and layout, while object-level generated meshes guide the agent toward detailed, compact geometry; articulated rigid objects are decomposed into movable parts with explicit joints. For deformables, category-wise agent sessions reconstruct curves as centerlines with radii, surfaces as manifold shells with thickness, and volumes as watertight solids for volumetric meshing. Reusable simulator skills initialize compatible physical models and parameters, while agent-guided behavioral tests expose mismatches and trigger targeted revisions of motion, geometry, numerics, or material modeling. On evaluated Replica and ScanNet++ scenes, CoDimRecon achieves competitive compositional reconstruction accuracy while additionally producing deformable assets for rod, shell, and solid simulation. We further demonstrate robot interactions across all three representations, including a controlled paper-folding case in which behavioral testing motivates plastic bending.
\end{abstract}

\section{Introduction}
Reconstructing simulation-ready indoor scenes from captured observations turns real environments into reusable digital worlds. Such scenes support applications ranging from robot learning~\citep{nasiriany2024robocasa} and computer games~\citep{li2024advances} to immersive content creation~\citep{luo2025vr}. Given a captured video or multi-view RGB images, we aim to recover a compositional 3D scene in which the room structure is explicit and each persistent object has its own geometry, pose, material, and physical model, with explicit articulation for movable rigid parts when applicable.

Existing compositional scene reconstruction methods are either optimization-based or zero-shot. Optimization-based approaches~\citep{wu2023objectsdf++,ni2024phyrecon,xia2026holoscene,ni2025decompositional,yang2025instascene,liu2024gaussian} optimize decomposed representations from multi-view images and masks, while zero-shot approaches compose pretrained reconstruction modules~\citep{xia2026simrecon,dong2026ras} or directly infer object-centric scenes~\citep{siddiqui2026shaper,wu2026jrm,xia2026fire3d}. However, these methods largely treat scene objects as rigid bodies, leaving deformable reconstruction and simulation underexplored.

Thin structures make this gap especially important. In physics-based simulation~\citep{li2026physics}, cables, cloth, paper, and other slender bodies are often not discretized as ordinary 3D solids: resolving a very small thickness volumetrically can require fine through-thickness resolution and, with standard low-order formulations, can suffer locking as structures become thinner~\citep{bischoff2004thinwalled}. Instead, decades of simulation research have developed codimensional models that represent rods as curves and shells as surfaces embedded in 3D~\citep{grinspun2003discreteshells,bergou2008discreteelasticrods}. Dimensional reduction avoids explicitly meshing thickness, but introduces additional bending and, for rods, twisting mechanics. It therefore changes the reconstruction target itself: a cable needs a continuous centerline and radius, a sheet a manifold midsurface and thickness, and a volumetric soft body a watertight solid suitable for volumetric meshing. Irreversible deformation adds another practical gap. Everyday interactions such as folding paper cannot be represented by elasticity alone; they require a plastic model that retains deformation after unloading. Existing compositional reconstruction pipelines with physical modeling largely focus on rigid-body physics and have not been designed around either codimensional geometry or such irreversible behavior.

Recent multimodal agents provide a natural interface between visual observations, 3D authoring tools, and physical simulators~\citep{gpt_astra}. We therefore investigate agentic reconstruction of simulation-ready scenes containing rigid objects, including articulated ones, alongside deformable objects. Direct agentic reconstruction from images alone can misjudge scale and layout or replace visible detail with coarse approximations. Deformables add a further challenge: their representation, physical model, and parameters cannot be determined from static geometry alone, and whether they support the intended interaction often becomes evident only when exercised in simulation.

To address these challenges, we introduce {\bf CoDimRecon}, an agentic framework for reconstructing simulation-ready scenes with rigid objects and deformable curves, surfaces, and volumes. Scene-level geometric priors ground scale and layout, while generated object meshes provide detailed shape references for rebuilding compact, editable geometry. For articulated rigid objects, the agent separates movable parts and assigns joints; for deformables, CoDimRecon reconstructs each geometric dimension separately: curves are completed into centerlines with radii, surfaces are repaired into manifold shells with thickness, and volumes are closed for tetrahedral meshing. Reusable simulator skills then initialize compatible physical models and parameters.

Finally, CoDimRecon uses agent-guided behavioral tests to check whether each deformable asset supports its intended interaction. The agent plans preset diagnostic manipulations, checks each rollout against physical acceptance criteria and expected behavior, and attributes failures to motion, geometry, numerics, or the material model. Material models are revised only when a behavioral test exposes a mismatch; for example, paper that springs back after folding triggers plastic hinge bending. This closes the loop from visual reconstruction to deformable assets represented and tested in the form required by downstream simulation.

Our contributions can be summarized as follows:
\begin{itemize}
    \item We present CoDimRecon, an agentic framework for reconstructing editable, simulation-ready indoor scenes from RGB observations, using complementary geometric and generative references to improve scene layout and object geometry.
    \item We extend simulation-ready reconstruction to deformables across geometric dimensions, producing solver-compatible curves, surfaces, and volumes for rod, shell, and solid simulation and initializing their physical models from reusable simulator skills.
    \item We close the reconstruction-to-simulation loop with agent-guided behavioral tests that trigger targeted revisions of motion, geometry, numerics, or material modeling. Paper folding provides a controlled elastic/plastic example, while the framework remains competitive on compositional reconstruction metrics across the evaluated Replica and ScanNet++ scenes.
\end{itemize}

\section{Related Work}

\leavevmode
\begin{wraptable}{r}{0.60\textwidth}
\vspace{-10pt}
\centering
\begingroup
\footnotesize
\setlength{\abovecaptionskip}{4pt}
\setlength{\tabcolsep}{1.6pt}
\renewcommand{\arraystretch}{1.05}
\resizebox{\linewidth}{!}{%
\begin{tabular}{@{}llcccccc@{}}
\toprule
Method & Input & \shortstack{Geometric\\Guidance} & \shortstack{Generative\\Prior} & \shortstack{Auto Instance\\Discovery} & \shortstack{Training-\\Free} & Agentic & \shortstack{Deformable\\Simulation} \\
\midrule
HoloScene & RGB, Mask, Cam & \cmark & \cmark & \xmark & \xmark & \xmark & \xmark \\
SimRecon & RGB & \cmark & \cmark & \cmark & \cmark & \xmark & \xmark \\
ReplicateAnyScene & RGB & \cmark & \cmark & \cmark & \cmark & \xmark & \xmark \\
VIGA & Single-view RGB & \xmark & \cmark & \cmark & \cmark & \cmark & \xmark \\
Lucida & RGB & \cmark & \cmark & \cmark & \xmark & \cmark & \xmark \\
Lumera & Single-view RGB & \cmark & \cmark & \cmark & \xmark & \cmark & \xmark \\
LiteReality-Agent & Posed RGBD & \cmark & \cmark & \cmark & \cmark & \cmark & \xmark \\
GPT-6 Astra & RGB & \xmark & \xmark & \cmark & \cmark & \cmark & \xmark \\
\midrule
\textbf{Our Method} & RGB & \cmark & \cmark & \cmark & \cmark & \cmark & \cmark \\
\bottomrule
\end{tabular}%
}
\caption{\textbf{Comparison of compositional scene-reconstruction methods.} Cam: camera parameters; RGBD: RGB images with depth.}
\label{tab:related_work_comparison}
\endgroup
\vspace{-8pt}
\end{wraptable}%
\noindent\textbf{Multi-View Compositional Scene Reconstruction.}~This
task reconstructs a scene as individually represented objects and their spatial arrangement from captured images or video. Existing approaches optimize object-level signed distance fields~\citep{wu2023objectsdf++,ni2024phyrecon,ni2025decompositional}, incorporate generative priors for partial-observation completion~\citep{yang2025instascene,xia2026holoscene,siddiqui2026shaper,wu2026jrm,xia2026fire3d}, retrieve reusable CAD assets~\citep{huang2025litereality,yu2025metascenes}, or assemble instances from pretrained 3D generators~\citep{xia2026simrecon,dong2026ras}. Concurrently, several agentic pipelines have emerged for captured-scene authoring~\citep{lucida2026,huang2026la,chen2026lumera}. As concurrent works, they are not directly comparable as empirical baselines: Lucida~\citep{lucida2026} and Lumera~\citep{chen2026lumera} rely on task-specific fine-tuning or RL policies for layout parsing and pose refinement, while LiteReality-Agent~\citep{huang2026la} requires privileged RGB-D scans and structural RoomPlan layouts rather than multi-view RGB alone.

Tab.~\ref{tab:related_work_comparison} compares these methods with ours. Among the listed methods, none reconstructs deformable curves, surfaces, and volumes for physical simulation or models plastic behavior. CoDimRecon additionally supports articulated rigid parts, uses generated meshes as references rather than final assets, and verifies reconstructed deformables through simulated robot interaction.

\noindent\textbf{Deformable and Codimensional Simulation.}~Physics-based simulation has developed mature models for rods, shells, volumetric solids, contact across mixed dimensions, and inelastic materials, but these methods generally assume solver-ready geometry and material models as input. Appendix~\ref{app:deformable_sim_related} reviews this literature in more detail; CoDimRecon targets the complementary problem of reconstructing such assets from visual observations.

\noindent\textbf{Single-Image Compositional Scene Reconstruction.}~Gen3DSR~\citep{ardelean2025gen3dsr}, SceneMaker~\citep{shi2026scenemaker}, and TabletopGen~\citep{wang2026tabletopgen} recover compositional scenes from a single image. VIGA~\citep{yin2026viga} reconstructs editable scene programs through a code--render--inspect loop from a single view; because it takes one image rather than multi-view observations of a specific room, its setting differs from ours and thus we do not evaluate against it. $\phi$-Scene~\citep{li2026phiscene}, REST3D~\citep{ma2026rest3d}, and SimuScene~\citep{lee2026simuscene} refine object layout through rigid-body physics. Our method instead uses multi-view geometric evidence and extends simulation-ready reconstruction to representation-specific deformable curves, surfaces, and volumes, with material models revised when required by the target behavior.

\noindent\textbf{Text-Driven Scene Synthesis.}~SceneSmith~\citep{pfaff2026scenesmith} and SAGE~\citep{xia2026sage} generate 3D environments from language or task specifications, while MUSE~\citep{xu2026muse} supports incremental construction and local editing through explicit requirements and verification. PAT3D~\citep{lin2026pat3d} uses differentiable rigid-body simulation to refine text-generated scene layouts, and GIF~\citep{xu2026gif} targets functional object compositions with geometric and physical guidance. GS-Agent~\citep{zhang2026gsagent} integrates a physics engine in an agentic loop to tune material parameters for text-driven 4D world generation. These methods synthesize scenes to satisfy user specifications; our task is to reconstruct the geometry, layout, and deformable objects of a particular observed environment, with deformable assets verified through simulated robot contact.

\section{Method}
\label{sec:method}

\begin{figure}[t]
    \centering
    \includegraphics[width=\textwidth]{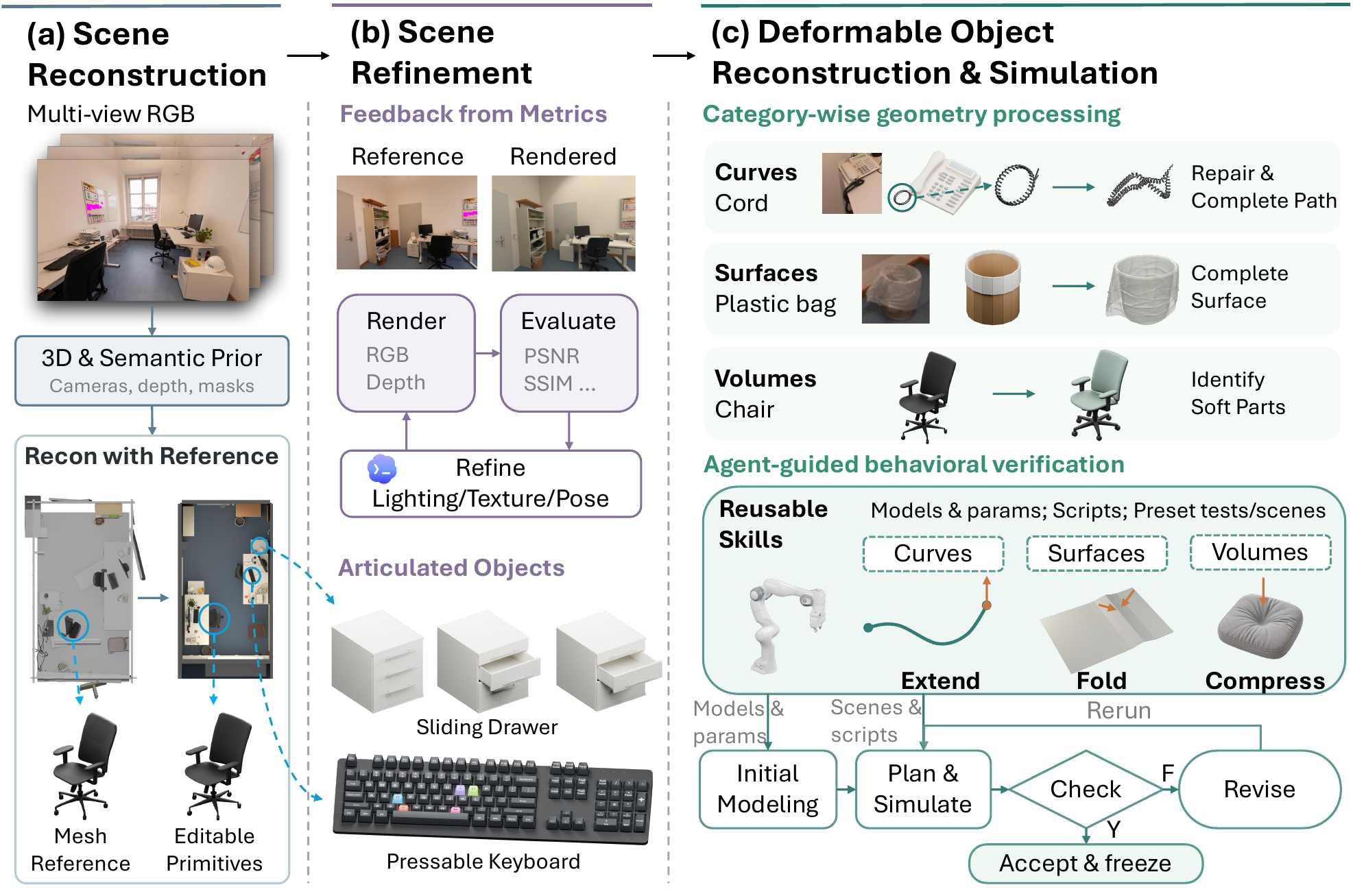}
    \vspace{-15pt}
    \caption{\textbf{Overview.}
    (a)~Geometric context and generated meshes guide editable primitive-based scene reconstruction.
    (b)~Render--evaluate--refine improves appearance and pose; articulated rigid objects receive joints, and rigid bodies are settled under gravity.
    (c)~Separate sessions reconstruct solver-compatible curves, surfaces, and volumes; reusable skills initialize physical models, and agent-guided robot tests diagnose motion, geometry, or numerical issues before material changes.}
    \label{fig:method_overview}
\end{figure}

As shown in Fig.~\ref{fig:method_overview}, CoDimRecon takes multi-view RGB observations $\mathcal{I}=\{I_v\}_{v=1}^{V}$ and proceeds in three stages. The agent first authors an editable scene using geometric context and generated meshes as references (Sec.~\ref{sec:method:enhance}), then refines appearance, articulation, and rigid-body stability (Sec.~\ref{sec:method:refine}). Finally, it reconstructs deformable curves, surfaces, and volumes and tests their behavior through simulated robot interaction (Sec.~\ref{sec:method:deform}).

\subsection{Reconstruction with Geometric and Generative References}
\label{sec:method:enhance}
In preliminary image-to-Blender trials, direct agentic reconstruction showed three recurring limitations: errors in scene scale and layout, coarse approximations of visible shape details, and limited use of external 3D tools even when available. We therefore make scene-level geometric context and object-level generated meshes explicit inputs to a fixed reconstruction workflow.

\noindent\textbf{Geometric Context.}~VGGT-Omega~\citep{wang2026vggtomega} provides camera intrinsics, camera-to-world poses, depth maps $D_v$, and back-projected point maps. The agent queries this shared metric context when estimating object dimensions and poses and when comparing renders with the observations, grounding object-level fitting in the scene layout.

\noindent\textbf{Obtaining Mesh References.}~Before agent authoring, we construct an object-level reference scene. CropFormer~\citep{qilu2023high} extracts entity masks per frame; we back-project and cluster them across views by 3D overlap following MaskClustering~\citep{yan2024maskclustering}, with InstaScene's under-segmentation filter~\citep{yang2025instascene}. For each 3D track, SAM3D~\citep{sam3dobj} generates a mesh from the most informative view, which is registered to the scene following ReplicateAnyScene~\citep{dong2026ras}. The agent retrieves the reference nearest to a target object's point cloud. Small missing objects are recovered with REST3D~\citep{ma2026rest3d} within their supporting-object regions. These meshes serve only as structural references; Appendices~\ref{sec:appx:gen_align} and~\ref{sec:appx:container_completion} give details, and Appendix~\ref{app:reimpl_ras} compares this design with 2D mask propagation.

\noindent\textbf{Workflow and Primitive Fitting.}~The agent loads the reference scene into Blender, removes redundant or erroneous objects, and refines the room envelope using the geometric context. For furniture and regular rigid objects, the generated mesh remains a shape reference rather than the final asset: the agent rebuilds the object from Blender primitives shaped with bevel, subdivision, and lattice modifiers. This yields compact, editable geometry, lets the agent correct implausible reference regions against the observations, and exposes parts for articulation. Deformable objects use the representations of Sec.~\ref{sec:method:deform}.

\subsection{Scene Refinement}
\label{sec:method:refine}

\textbf{Rendering Enhancement with Metric Evaluation.}~Agent-authored scenes often differ from the observations in lighting and appearance, so the agent runs a render--evaluate--refine loop with quadratic color alignment~\citep{zhang2025real} as a diagnostic. Using iteratively reweighted least squares, it fits a per-channel mapping $c' = a\,c^{2} + b\,c + d$ from rendered to reference pixel values. A large improvement after alignment suggests a photometric mismatch in lighting or materials; little improvement directs attention to geometry or pose. The agent edits the corresponding scene components and reevaluates the input views. We keep these edits explicit rather than baking a 3DGS appearance onto the mesh, which can entangle illumination with albedo and become inconsistent under simulation lighting.

\noindent\textbf{Articulated Object Reconstruction.}~We provide a joint-modeling guide covering revolute, prismatic, screw, cylindrical, universal, and spherical joints. The agent separates independently movable rigid parts and assigns explicit joints, including fine components such as keyboard keys and telephone buttons when applicable. This decomposition also exposes geometry that might otherwise be collapsed into a coarse textured proxy.

\noindent\textbf{Rigid-Body Stabilization.}~Small pose errors can leave objects floating or interpenetrating and destabilize downstream simulation. We therefore import the scene into MuJoCo~\citep{todorov2012mujoco}, treat all objects as rigid at this stage, settle them under gravity, and write the resulting poses back as the canonical placements. Deformable simulation is handled separately in Sec.~\ref{sec:method:deform}.

\subsection{Category-wise Deformable Object Reconstruction and Simulation}
\label{sec:method:deform}

Among the compositional reconstruction methods in Tab.~\ref{tab:related_work_comparison}, none reconstructs deformable curves, surfaces, and volumes for physical simulation. We therefore reconstruct deformables in representation-specific forms---curves, surfaces, and volumes---and configure them for downstream simulation via agent-guided behavioral verification.

\noindent\textbf{Deformable Object Reconstruction.}~Simulation requires geometry compatible with its discretization. The required repairs depend on geometric dimension: curves such as cables need path completion and radius estimation; surfaces such as clothing and paper need hole patching and fragment merging into a manifold shell with thickness; and volumes such as cushions need watertight closure for volumetric meshing. We preserve this dimensionality in simulation, using rod, shell, and solid discretizations for curves, surfaces, and volumes, respectively. We process the three categories in separate agent sessions; a cable ablation shows that combining all instructions degrades curve reconstruction (Fig.~\ref{fig:aba_cate_wise}).

\begin{table*}[t]
\centering
\small
\setlength{\tabcolsep}{4pt}
\renewcommand{\arraystretch}{1.15}
\resizebox{\textwidth}{!}{%
\begin{tabular}{@{}llccccccc@{}}
\toprule
Object & Model & $\rho$ & $E$ & $\nu$ & $r/h$ & $\kappa_Y$ & $\delta$ & $\mu$ \\
& & $(\mathrm{kg/m^3})$ & $(\mathrm{MPa})$ & & $(\mathrm{mm})$ & $(\mathrm{m^{-1}})$ & $(\mathrm{mm})$ & \\
\midrule
Telephone cord & Discrete elastic rod               & 1600  & 40   & 0.35 & 1.3   & -- & 0.15 & 0.60 \\
Paper          & StVK membrane, plastic hinges      & 761.9 & 2250 & 0.15 & 0.105 & 25 & 0.50 & 0.40 \\
Chair cushion  & StVK--Hencky solid                 & 100   & 0.03 & 0.20 & --    & -- & 0.50 & 0.40 \\
Beanbag        & Stable Neo-Hookean solid           & 120   & 0.02 & 0.30 & --    & -- & 0.50 & 0.40 \\
\bottomrule
\end{tabular}%
}
\caption{\textbf{Deformable simulation parameters.}
$\rho$: density; $E$: Young's modulus (paper: membrane and bending; cord: stretching, bending, and twisting);
$\nu$: Poisson's ratio; $r/h$: cord radius or shell thickness;
$\kappa_Y$: plastic-hinge yield curvature; $\delta$: contact activation distance;
$\mu$: friction coefficient. Dashes denote inapplicable entries.
The cord's stress-free natural shape is the reconstructed coil, so the rod keeps its coiled shape without load.
Values are effective simulation parameters, not measured material properties; paper plasticity and the cushion's StVK--Hencky model were selected through behavioral testing.
Appendix~\ref{sec:appx:deformable_settings} gives discretization, boundary conditions, and tasks.}
\label{tab:material_parameters}
\end{table*}

\noindent\textbf{Physical Modeling from Simulator Examples.}~Reconstructed geometry does not determine the material and contact properties needed for simulation. Our backend uses IPC-family contact~\citep{li2020ipc}, while reusable skills package available constitutive models, runnable example scenes, scripts, and candidate initial parameters. The geometric representation determines the discretization---rod for curves, shell for surfaces, and solid finite elements for volumes---while the agent selects an asset-appropriate material model and adapts the closest reference configuration. Thus the telephone cord uses a discrete elastic rod, paper uses a shell model with plastic hinges, and the cushions use solid finite elements. Tab.~\ref{tab:material_parameters} summarizes the resulting parameters, and Appendix~\ref{sec:appx:deformable_settings} gives the discretizations.

\noindent\textbf{Rest Shape.}~Reconstructed deformable geometry is not necessarily in equilibrium under gravity, so each asset is settled before interaction. Curves and surfaces are imported directly; if settling produces excessive deformation, the agent revises the geometry and repeats the process. For high-polygon volumes, we instead build a closed low-poly proxy, bind the detailed mesh to it, tetrahedralize the proxy, and settle that representation. The resulting rest shape is written back to Blender while preserving the simulation parameters and editable asset.

\noindent\textbf{Agent-Guided Behavioral Verification.}~Simulator examples provide only an initialization, so we use behavioral tests to check whether each asset supports its intended interaction. The agent plans preset diagnostic manipulations rather than requiring real-robot rollouts for calibration~\citep{zhang2025real}. The robot lifts a telephone handset to extend the cord, presses and releases a chair cushion at three locations, and folds paper that should retain a crease after release. Each task has physical validity checks, such as bounds on element stretch and inversion, together with task-level acceptance criteria. The agent specifies grasp and approach poses, gripper actuation, and arm motion; the simulator executes the task offline, and the agent reviews the rendered rollout and diagnostic audit.

When a rollout fails, the agent diagnoses the failure in a fixed order: commanded motion, tool, and contact location; geometry and boundary conditions; then numerical settings. The material model changes only if the required behavior still cannot be reproduced, after which it remains fixed for that asset. For paper, springback after release triggers plastic hinge bending; replaying the same trajectory with plasticity disabled isolates its effect on crease retention (Fig.~\ref{fig:paper_fold_verification}, Tab.~\ref{tab:paper_fold}). Because no measured real dynamics are available, accepted values are effective properties under the assumed model, not recovered material parameters. Appendix~\ref{sec:appx:deformable_settings} lists the tasks and acceptance criteria, and Tab.~\ref{tab:revision_log} records each revision.

\section{Experiments}
\label{sec:exp}

\subsection{Setup}
\noindent\textbf{Datasets.} We evaluate three Replica~\citep{straub2019replica} scenes and three ScanNet++~\citep{yeshwanth2023scannetpp} scenes from the HoloScene~\citep{xia2026holoscene} release, covering varied indoor layouts, object density, and lighting.

\noindent\textbf{Quantitative Baselines.} \textbf{1) HoloScene$^{*}$}~\citep{xia2026holoscene} denotes our modified configuration of this optimization-based method for simulation-ready 3D worlds. We run its official code with normal supervision disabled, using our mask-clustering instance masks and VGGT-Omega camera poses and depth. The asterisk identifies results specific to this modified configuration. \textbf{2) ReplicateAnyScene}~\citep{dong2026ras} is a zero-shot compositional pipeline; we use its default VLM~+~SAM3 segmentation and reimplement the unreleased pose-alignment and relation-reasoning modules. \textbf{3) GPT-6 Astra}~\citep{gpt_astra} is an agent-only baseline that directly authors each scene in Blender from the RGB frames with reasoning effort \texttt{xhigh}. SimRecon~\citep{xia2026simrecon} appears in the capability comparison (Tab.~\ref{tab:related_work_comparison}) but not the quantitative benchmark because its object-completion and FoundationPose-based object-pose refinement modules were not publicly available at evaluation time. Appendix~\ref{app:reimpl} details baseline implementations, and Appendix~\ref{app:prompt_gpt6_astra} gives the GPT-6 Astra prompt.

\noindent\textbf{Metrics.} Geometry is evaluated with Chamfer distance (CD, cm), F1$@5$\,cm, and normal consistency (NC). Rendering uses PSNR, SSIM, and LPIPS on the same input views used for reconstruction, so these scores measure observation fidelity rather than novel-view synthesis. Latent Similarity~\citep{tang2026bvb} compares matched source and rendered clips in a frozen V-JEPA~2.1 encoder~\citep{murlabadia2026vjepa21unlockingdense}, with Layout and Motion as feature-space similarity measures. As a rigid-body stability proxy, we release each dynamic object individually in MuJoCo while all others remain fixed and report the fraction that stay in place: Stable(Ground) for floor-contact objects and Stable(All) for all dynamic objects. Appendix~\ref{app:eval_details} gives the full protocols.

\subsection{Results}

\begin{table*}[t]
\resizebox{\textwidth}{!}{
\small
\centering
\begin{tabular}{llcccccccccc}
\toprule
\multirow{3}{*}{Datasets} & \multirow{3}{*}{Method} & \multicolumn{3}{c}{Geometry} & \multicolumn{3}{c}{Rendering} & \multicolumn{2}{c}{Latent Similarity} & \multicolumn{2}{c}{Rigid Stability} \\
\cmidrule(lr){3-5} \cmidrule(lr){6-8} \cmidrule(lr){9-10} \cmidrule(lr){11-12}
& & \multirow{2}{*}{CD$\downarrow$} & \multirow{2}{*}{F1$\uparrow$} & \multirow{2}{*}{NC$\uparrow$} & \multirow{2}{*}{PSNR$\uparrow$} & \multirow{2}{*}{SSIM$\uparrow$} & \multirow{2}{*}{LPIPS$\downarrow$} & \multirow{2}{*}{Layout$\uparrow$} & \multirow{2}{*}{Motion$\uparrow$} & Stable & Stable \\
& & & & & & & & & & (Ground) $\%$ $\uparrow$ & (All) $\%$ $\uparrow$ \\
\midrule
\multirow{4}{*}{Replica}
& HoloScene$^{*}$ & \cellcolor{gsecond}6.79 & \cellcolor{gbest}53.48 & \cellcolor{gsecond}78.76 & \cellcolor{gbest}21.22 & \cellcolor{gbest}0.6881 & \cellcolor{gbest}0.4171 & 94.46 & \cellcolor{gsecond}95.71 & 77.77 & 49.41 \\
& ReplicateAnyScene & 41.88 & 18.74 & 61.82 & 11.17 & 0.5529 & 0.6461 & 76.84 & 80.88 & \cellcolor{gsecond}96.97 & 69.47 \\
& GPT-6 Astra & 11.79 & \cellcolor{gsecond}52.91 & 73.12 & 14.07 & 0.5671 & 0.5440 & \cellcolor{gbest}97.24 & 95.57 & 92.27 & \cellcolor{gsecond}87.97 \\
& Our Method & \cellcolor{gbest}6.63 & 52.24 & \cellcolor{gbest}79.76 & \cellcolor{gsecond}15.06 & \cellcolor{gsecond}0.5695 & \cellcolor{gsecond}0.4817 & \cellcolor{gsecond}95.14 & \cellcolor{gbest}96.90 & \cellcolor{gbest}100.00 & \cellcolor{gbest}97.54 \\
\cmidrule(lr){1-12}
\multirow{4}{*}{ScanNet++}
& HoloScene$^{*}$ & 27.15 & 28.49 & 65.24 & \cellcolor{gbest}18.79 & \cellcolor{gbest}0.7134 & \cellcolor{gsecond}0.3959 & 80.20 & 81.73 & 55.40 & 42.03 \\
& ReplicateAnyScene & 50.73 & 19.08 & 54.69 & 8.94 & 0.5483 & 0.6710 & 92.87 & 74.87 & 78.66 & 82.72\\
& GPT-6 Astra & \cellcolor{gsecond}22.73 & \cellcolor{gsecond}35.90 & \cellcolor{gsecond}70.14 & 14.22 & 0.5579 & 0.5356 & \cellcolor{gsecond}94.47 & \cellcolor{gsecond}96.31 & \cellcolor{gsecond}96.30 & \cellcolor{gbest}96.31 \\
& Our Method & \cellcolor{gbest}21.66 & \cellcolor{gbest}38.28 & \cellcolor{gbest}74.91 & \cellcolor{gsecond}16.68 & \cellcolor{gsecond}0.6440 & \cellcolor{gbest}0.3728 & \cellcolor{gbest}96.30 & \cellcolor{gbest}97.04 & \cellcolor{gbest}97.05 & \cellcolor{gbest}96.31 \\
\bottomrule
\end{tabular}
}
\caption{\textbf{Compositional scene-reconstruction results.}
CoDimRecon produces the most physically stable reconstructions, with the lowest CD and highest NC on both datasets, while keeping input-view rendering second only to HoloScene$^{*}$ in PSNR and SSIM. HoloScene$^{*}$ uses a modified configuration with normal supervision disabled, our mask-clustering instance masks, and VGGT-Omega camera poses and depth (Appendix~\ref{app:reimpl_holoscene}). Dark and light blue mark the best and second-best results; ties share the same color.}
\label{tab:main_results}
\vspace{-1em}
\end{table*}

\begin{figure}[t]
    \centering
    \includegraphics[width=\textwidth]{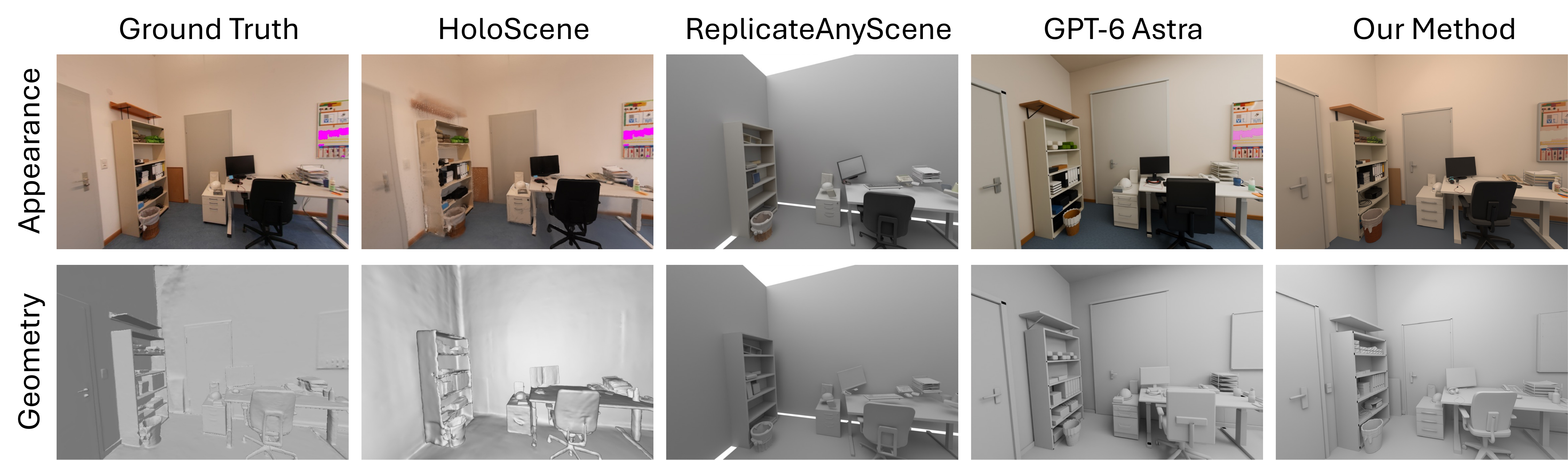}
    \caption{\textbf{Qualitative compositional scene-reconstruction comparison.} We compare rendered appearance and geometry on ScanNet++ scene \texttt{67d702f2e8}. To reduce the influence of upstream camera-pose and instance-segmentation errors, we show HoloScene and ReplicateAnyScene with SfM-estimated camera poses from ScanNet++ and manually annotated instance masks from the HoloScene release. The HoloScene result comes from its official release, and ReplicateAnyScene comes from our reimplementation. These baseline configurations differ from those in Tab.~\ref{tab:main_results}, where HoloScene$^{*}$ uses our estimated inputs with normal supervision disabled. GPT-6 Astra reconstructs directly from RGB images. Our method uses camera poses, depth, and instance masks estimated from multi-view RGB images.}
    \label{fig:scene_recon_vis}
\end{figure}

\noindent\textbf{Comparison with Baselines.}~Tab.~\ref{tab:main_results} reports scene-level results. Among the evaluated methods, ours achieves the lowest CD, the highest NC, and the best rigid-body stability on both datasets, while ranking second in input-view PSNR and SSIM. On Replica, the CD and NC gains over the strongest baseline are modest, and our F1 trails the top two methods. GPT-6 Astra shares our agent backbone but reconstructs directly from RGB; Sec.~\ref{sec:exp:ablation} studies the effect of our reference inputs. Fig.~\ref{fig:scene_recon_vis} provides a complementary qualitative comparison using dataset-provided poses and annotated masks for HoloScene and ReplicateAnyScene, with configurations detailed in the caption. In this comparison, the baselines exhibit varying degrees of incomplete geometry, missing scene elements, and scale inconsistencies, while our method better preserves object details and scene completeness. PSNR and SSIM compare pixels, so they penalize the lighting and material differences that remain in our explicitly authored scenes (Sec.~\ref{sec:method:refine}). Layout and Motion instead compare source and rendered clips in the feature space of a frozen video encoder. Our method achieves the highest Motion similarity on both datasets and the highest Layout similarity on ScanNet++, complementing the qualitative evidence for scene completeness and spatial fidelity. Because these feature-space scores also depend on rendering style, we use them as complementary evidence (Appendix~\ref{app:eval_latent}).

\begin{figure}[t]
    \centering
    \begin{minipage}[t]{0.5\textwidth}
        \vspace{0pt}
        \centering
        \setlength{\parskip}{0pt}
        \captionsetup[subfloat]{farskip=2pt,nearskip=0pt,captionskip=2pt}
        \subfloat[Telephone cord.\label{fig:cord_sim}]{%
            \includegraphics[width=\linewidth]{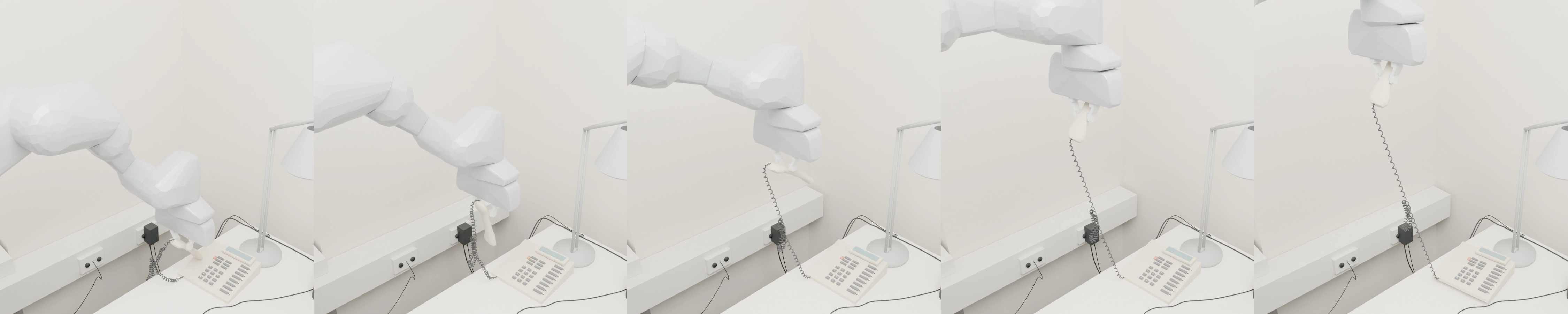}}
        \par
        \subfloat[Paper.\label{fig:paper_sim_detail}]{%
            \includegraphics[width=\linewidth]{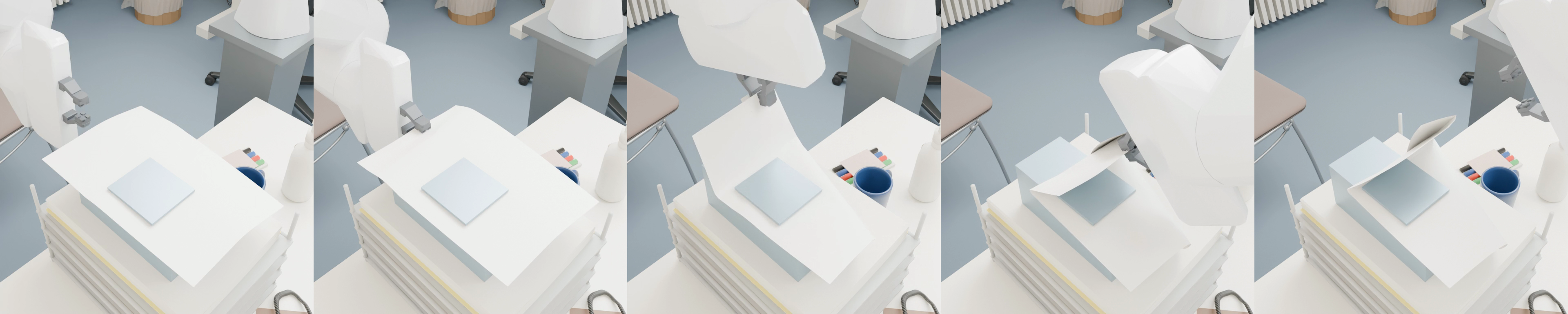}}
        \par
        \subfloat[Cushion.\label{fig:cushion_sim}]{%
            \includegraphics[width=\linewidth]{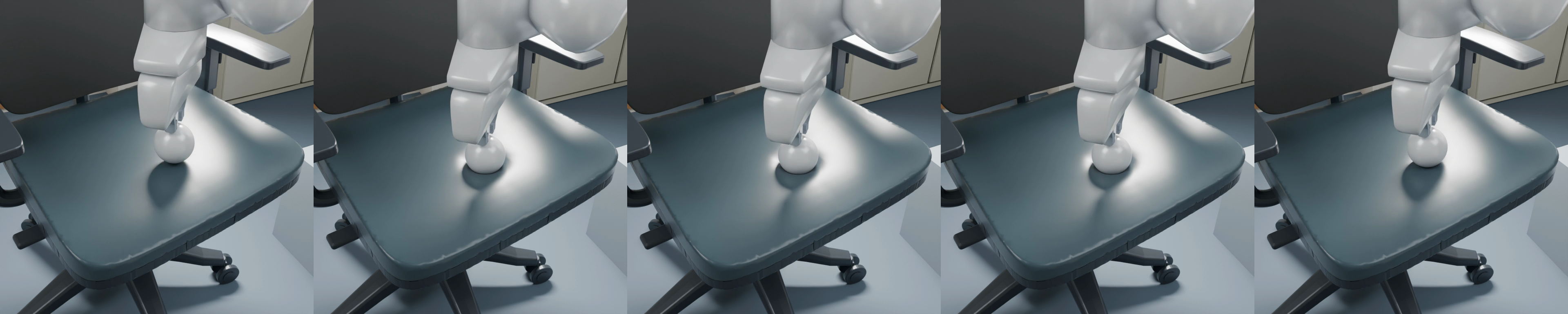}}
        \vspace{-10pt}
        \caption{\textbf{Deformable simulation examples.} Robot interactions with a reconstructed telephone cord (curve), paper (surface), and chair cushion (volume), shown left to right.}
        \label{fig:paper_sim}
    \end{minipage}\hfill
    \begin{minipage}[t]{0.49\textwidth}
        \vspace{0pt}
        \centering
        \includegraphics[width=\linewidth]{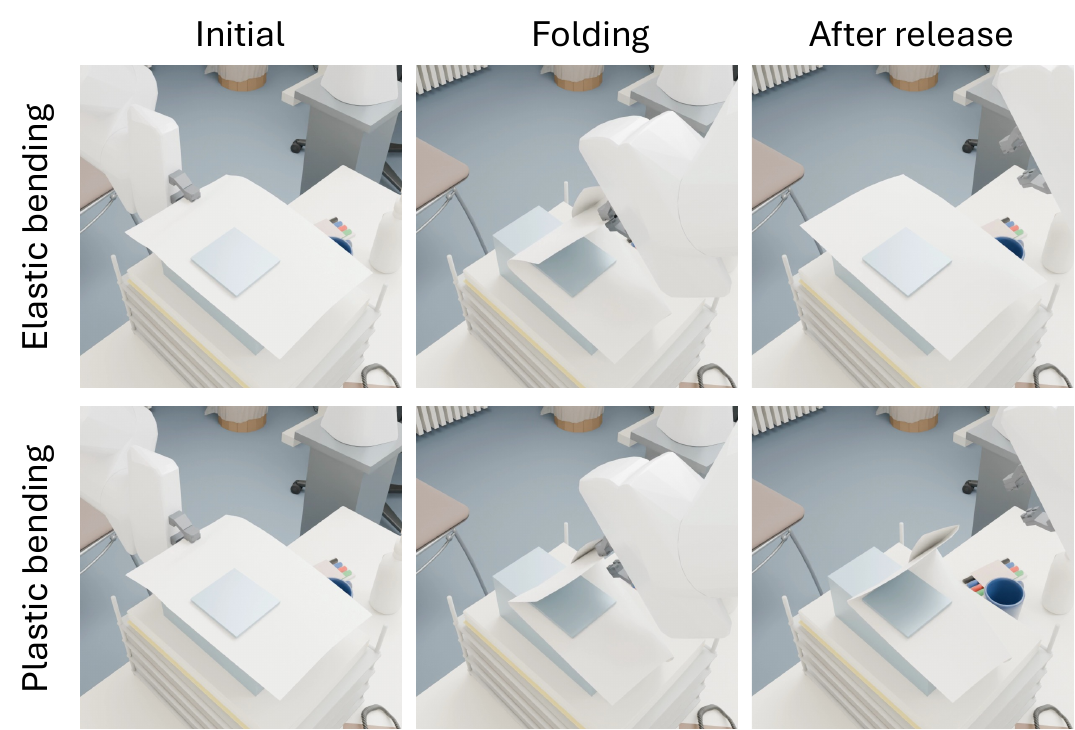}
        \caption{\textbf{Paper-folding behavioral test.} Under the same robot trajectory, elastic bending (top) springs back after release, while plastic bending (bottom) retains the fold. The elastic case is a controlled rerun with plasticity disabled.}
        \label{fig:paper_fold_verification}
    \end{minipage}
    \par\medskip
    \begin{minipage}{\textwidth}
        \centering
        {\small
        \begin{tabular}{lccccc}
        \toprule
        \multirow{2}{*}{Bending} & \multicolumn{3}{c}{Fold angle ($^\circ$)} & \multirow{2}{*}{Retained (\%)} & \multirow{2}{*}{Yielded} \\
        \cmidrule(lr){2-4}
         & Peak & $+1$\,s & End & & \\
        \midrule
        Elastic & 121.66 & 4.48   & 2.67   & 3.7  & 0   \\
        Plastic & 124.24 & 114.97 & 114.15 & 92.5 & 372 \\
        \bottomrule
        \end{tabular}\par}
        \captionof{table}{\textbf{Paper-folding comparison.} $+1$\,s: one second after opening; Retained: $+1$\,s angle divided by peak angle; Yielded: hinges whose natural angle changes by more than $0.1$\,rad. See Appendix~\ref{sec:appx:deformable_settings} for settings and angle definition.}
        \label{tab:paper_fold}
    \end{minipage}
\end{figure}

\noindent\textbf{Deformable Object Simulation Results.} Among the evaluated reconstruction baselines, none outputs deformable assets for physical simulation (Tab.~\ref{tab:related_work_comparison}), so we report capability demonstrations rather than a cross-method quantitative comparison. We simulate reconstructed curve, surface, and volume assets with an inserted robot and replay the trajectories in Blender for rendering. The telephone cord extends as the handset is raised (Fig.~\ref{fig:cord_sim}), the paper retains its crease after release (Fig.~\ref{fig:paper_sim_detail}), and the cushion indents under three presses and recovers (Fig.~\ref{fig:cushion_sim}). To make the indentation visible, the cushion is rendered with modified surface material and lighting; the simulation is unaffected. Demos of the reconstructed scenes and deformable objects are available on the project page (Appendix~\ref{app:supp_demo}).

\noindent\textbf{Behavioral Testing and Revision: Paper.}~The first folding rollout passed the physical-validity checks but sprang back after release, exposing a behavioral mismatch. The agent therefore enabled plastic hinge bending with yield curvature $25\,\mathrm{m^{-1}}$ and no hardening. To isolate this model change, we replay the same $140^\circ$ gripper trajectory with plasticity enabled or disabled while holding the remaining parameters fixed (Fig.~\ref{fig:paper_fold_verification}); the elastic control is a subsequent rerun rather than the original trial. One second after opening, the fold angle is $4.48^\circ$ with elastic bending and $114.97^\circ$ with plastic bending (Tab.~\ref{tab:paper_fold}), while both rollouts remain physically valid. This controlled case illustrates how behavioral testing can expose a material-model mismatch not detected by numerical validity checks. Tab.~\ref{tab:revision_log} records the revisions across assets.

The rest-shape settling study on a reconstructed beanbag is reported in Appendix~\ref{sec:appx:deformable_settings} (Fig.~\ref{fig:rest_shape_settling}).

\subsection{Ablation Study}
\label{sec:exp:ablation}

\begin{figure}[t]
    \centering
    \begin{minipage}[t]{0.43\textwidth}
        \vspace{0pt}
        \centering
        \includegraphics[width=\linewidth]{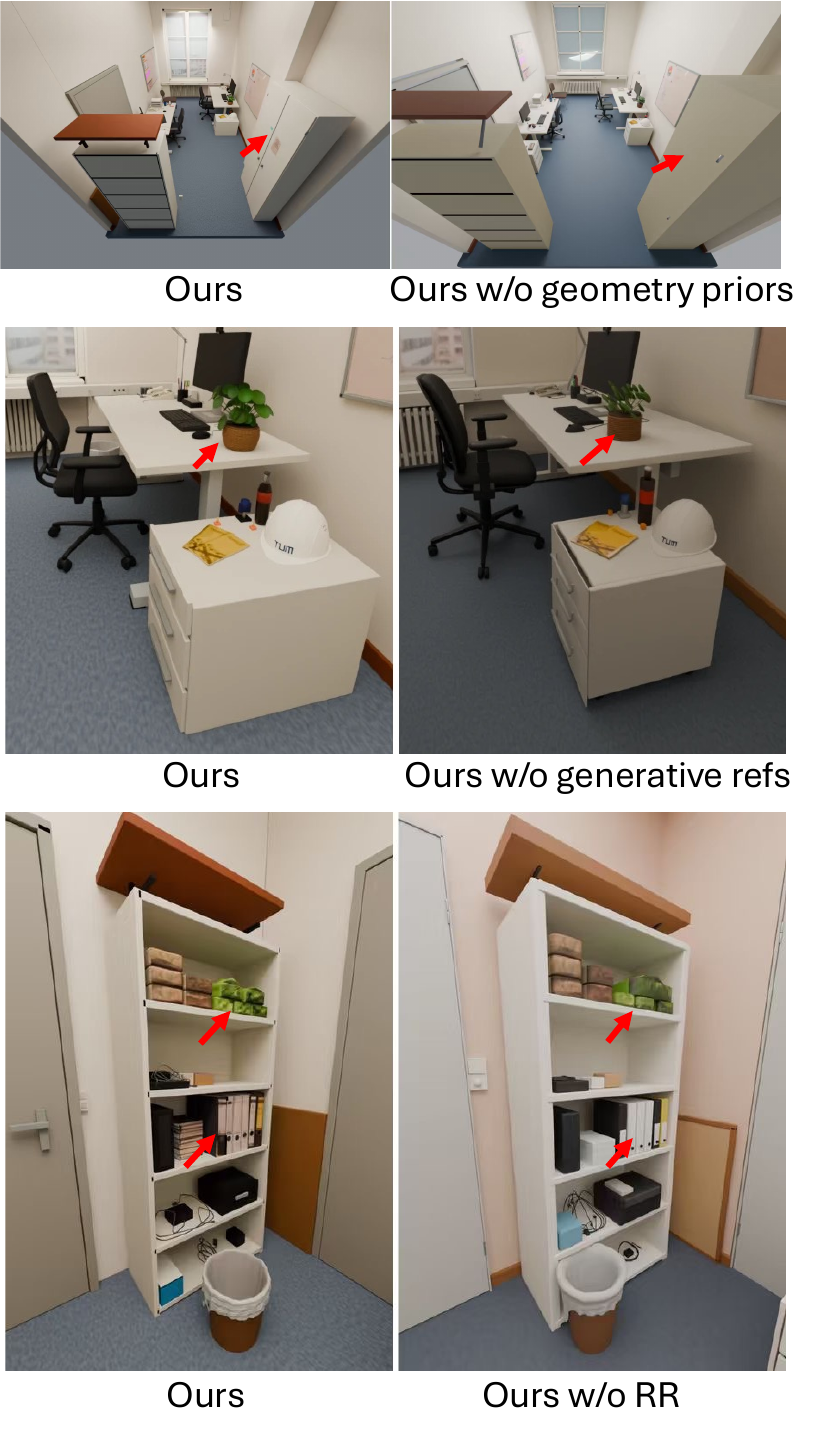}
    \end{minipage}\hfill
    \begin{minipage}[t]{0.53\textwidth}
        \vspace{0pt}
        \centering
        {\footnotesize
        \setlength{\tabcolsep}{3pt}
        \begin{tabular*}{\linewidth}{@{\extracolsep{\fill}}lccc@{}}
        \toprule
        \multicolumn{4}{@{}l}{\textbf{Geometry}} \\
        Variant & CD$\downarrow$ & F1$\uparrow$ & NC$\uparrow$ \\
        \midrule
        Full                & 7.24  & 44.98 & 75.93 \\
        w/o geometric refs  & 11.94 & 44.53 & 69.70 \\
        w/o generative refs & 9.86  & 39.12 & 68.14 \\
        w/o RR              & 7.36  & 43.72 & 75.41 \\
        \midrule
        \multicolumn{4}{@{}l}{\textbf{Rendering}} \\
        Variant & PSNR$\uparrow$ & SSIM$\uparrow$ & LPIPS$\downarrow$ \\
        \midrule
        Full                & 16.68 & 0.6440 & 0.3728 \\
        w/o RR              & 15.82 & 0.6343 & 0.3938 \\
        \bottomrule
        \end{tabular*}\par}
        \captionof{table}{\textbf{Component ablation.} Geometry on ScanNet++ scene \texttt{67d702f2e8}; rendering averaged over the evaluated ScanNet++ scenes. RR: rendering refinement.}
        \label{tab:ablation}
        \par\medskip
        \includegraphics[width=\linewidth]{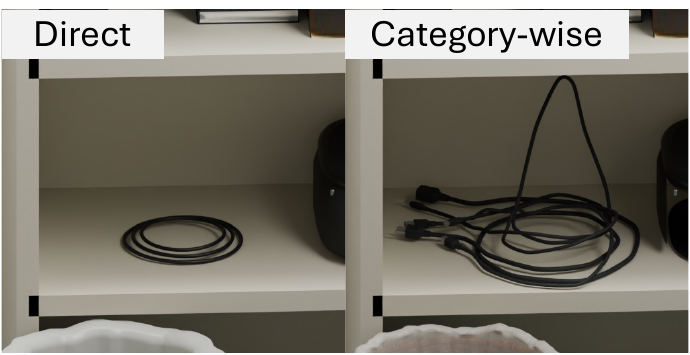}
    \end{minipage}
    \par
    \begin{minipage}[t]{0.43\textwidth}
        \caption{\textbf{Qualitative component ablation.} Each row compares the full method with one component removed; red arrows mark affected regions.}
        \label{fig:ablation}
    \end{minipage}\hfill
    \begin{minipage}[t]{0.53\textwidth}
        \captionof{figure}{\textbf{Separate-session ablation on cables.} Combined deformable-category instructions (left) produce coarser cable geometry than a dedicated curve session (right).}
        \label{fig:aba_cate_wise}
    \end{minipage}
\end{figure}

\noindent\textbf{Reference Priors.}~We remove the geometric and generative references one at a time and measure geometry on ScanNet++ scene \texttt{67d702f2e8} (Tab.~\ref{tab:ablation}, top); Fig.~\ref{fig:ablation} shows representative visual differences.
Without geometric references, CD increases by $64.9\%$, the largest increase among the geometry variants, and NC decreases by $8.2\%$, whereas F1 changes little. These changes are consistent with the scale and layout errors visible in Fig.~\ref{fig:ablation}, such as oversized cabinets.
Without generative references, F1 and NC show the largest drops among the geometry variants, decreasing by $13.0\%$ and $10.3\%$ relative to the full method, consistent with coarser object shapes when generated meshes are unavailable as references.
Because this geometry ablation covers a single scene, we read these differences as indicating the role of each reference rather than as benchmark-level gains.

\noindent\textbf{Rendering Refinement.}~As shown in the bottom of Tab.~\ref{tab:ablation}, we evaluate the render--evaluate--refine loop of Sec.~\ref{sec:method:refine} over the evaluated ScanNet++ scenes.
Without rendering refinement (RR), PSNR and SSIM decrease by $5.2\%$ and $1.5\%$, and LPIPS increases by $5.6\%$ relative to the full method.
On the geometry-ablation scene, the same variant increases CD by only $1.7\%$, so in this ablation the loop mainly improves appearance rather than geometry.

\noindent\textbf{Deformable-Category Instructions.}~In Sec.~\ref{sec:method:deform}, we reconstruct curves, surfaces, and volumes in separate agent sessions. To test this choice, we compare a single session that receives the instructions for all three categories with a dedicated curve session. As shown in Fig.~\ref{fig:aba_cate_wise}, the combined session produces coarser and less plausible cable geometry than the dedicated session. 

\section{Conclusion}
We presented CoDimRecon, an agentic framework for reconstructing editable, simulation-ready scenes with rigid objects and deformable curves, surfaces, and volumes. Geometric and generative references guide scene authoring, while representation-specific reconstruction produces solver-compatible assets for rod, shell, and solid simulation. Agent-guided behavioral tests then expose mismatches and trigger targeted revisions; paper folding provides a controlled example in which an elastic model passes numerical-validity checks yet requires plastic bending to retain the fold. Across the evaluated Replica and ScanNet++ scenes, CoDimRecon remains competitive on compositional reconstruction metrics while extending the output to simulation-ready deformables.

\subsubsection*{Acknowledgments}
We thank Yi-Ling Qiao, Xinyu Lu, and Kemeng Huang for their guidance on using the IPC-based simulator. This work is supported in part by the National Natural Science Foundation of China (Grant Nos.~92467204 and 62472249), the Shenzhen Science and Technology Program (Grant No.~KJZD20240903102300001), and gift funding from Genesis AI. Shuzhao Xie's work is supported by the Google Cloud Research Credits program. Shuzhao Xie thanks Chen Tang for help with computational resources.

\bibliography{arxiv_conference}
\bibliographystyle{arxiv_conference}

\appendix

\section{Additional Related Work: Deformable and Codimensional Simulation}
\label{app:deformable_sim_related}

\noindent\textbf{Thin structures and codimensional mechanics.}
Physics-based graphics has long treated thin structures with reduced-dimensional models rather than resolving their full thickness volumetrically; see~\citet{li2026physics} for a broader modern overview. Classical cloth simulation represents fabric as a triangulated surface with in-plane and bending response~\citep{baraff1998cloth}, while discrete-shell models formulate bending directly on triangle meshes and can represent sharp folds and changes in rest curvature~\citep{grinspun2003discreteshells}. Discrete elastic rods similarly reduce slender bodies to centerlines with bending and twisting energies~\citep{bergou2008discreteelasticrods}. These representations avoid the high through-thickness resolution and locking issues that can arise when very thin structures are treated as ordinary low-order 3D solids~\citep{bischoff2004thinwalled}. Recent work continues to improve thin-structure discretization, for example with smooth B-spline finite elements for cloth~\citep{meng2026bspline}. These methods make clear that the geometry required by a simulator depends on the object's effective dimension: rods need centerlines and radii, shells need midsurfaces and thicknesses, and solids need volumetric domains.

\noindent\textbf{Contact across dimensions.}
Thin structures also require robust contact handling. Classical cloth work developed collision and friction treatment for surface meshes~\citep{bridson2002cloth}; IPC later introduced intersection- and inversion-free variational contact~\citep{li2020ipc}, and C-IPC extended it to mixed-dimensional particles, rods, shells, and solids~\citep{li2021cipc}. Our backend uses this class of contact methods, while our contribution is the reconstruction of solver-compatible assets rather than a new contact formulation.

\noindent\textbf{Elastoplastic and inelastic behavior.}
Many real objects do not return to their original rest state after manipulation. Plasticity has therefore been modeled across several simulation paradigms. Discrete shells can encode permanent folds by changing rest dihedral angles~\citep{grinspun2003discreteshells}; elastoplastic constitutive laws have also been central to volumetric and particle methods, for example in snow simulation~\citep{stomakhin2013snow}. Energetically Consistent Inelasticity formulates finite-strain elastoplasticity and viscoelasticity for optimization-based FEM and MPM time integration~\citep{li2022inelasticity}. These works provide increasingly general tools for irreversible deformation; CoDimRecon uses this modeling capacity when a behavioral test requires it, as in the paper-folding example where elastic bending cannot retain a crease.

\noindent\textbf{Relation to simulation-ready reconstruction.}
The works above start from a prescribed rest geometry, discretization, constitutive model, and usually material parameters. In contrast, existing simulation-ready scene reconstruction has largely focused on rigid geometry and rigid-body physics. CoDimRecon bridges these areas by reconstructing representation-specific geometry for deformable curves, surfaces, and volumes, initializing compatible physical models from simulator examples, and revising them when behavioral tests expose a mismatch. Because static RGB observations do not identify true dynamic material parameters, the resulting values are treated as effective simulation parameters rather than measured material properties.

\section{Method Details}
\label{app:method_details}

\begin{figure}[h]
    \centering
    \includegraphics[width=\textwidth]{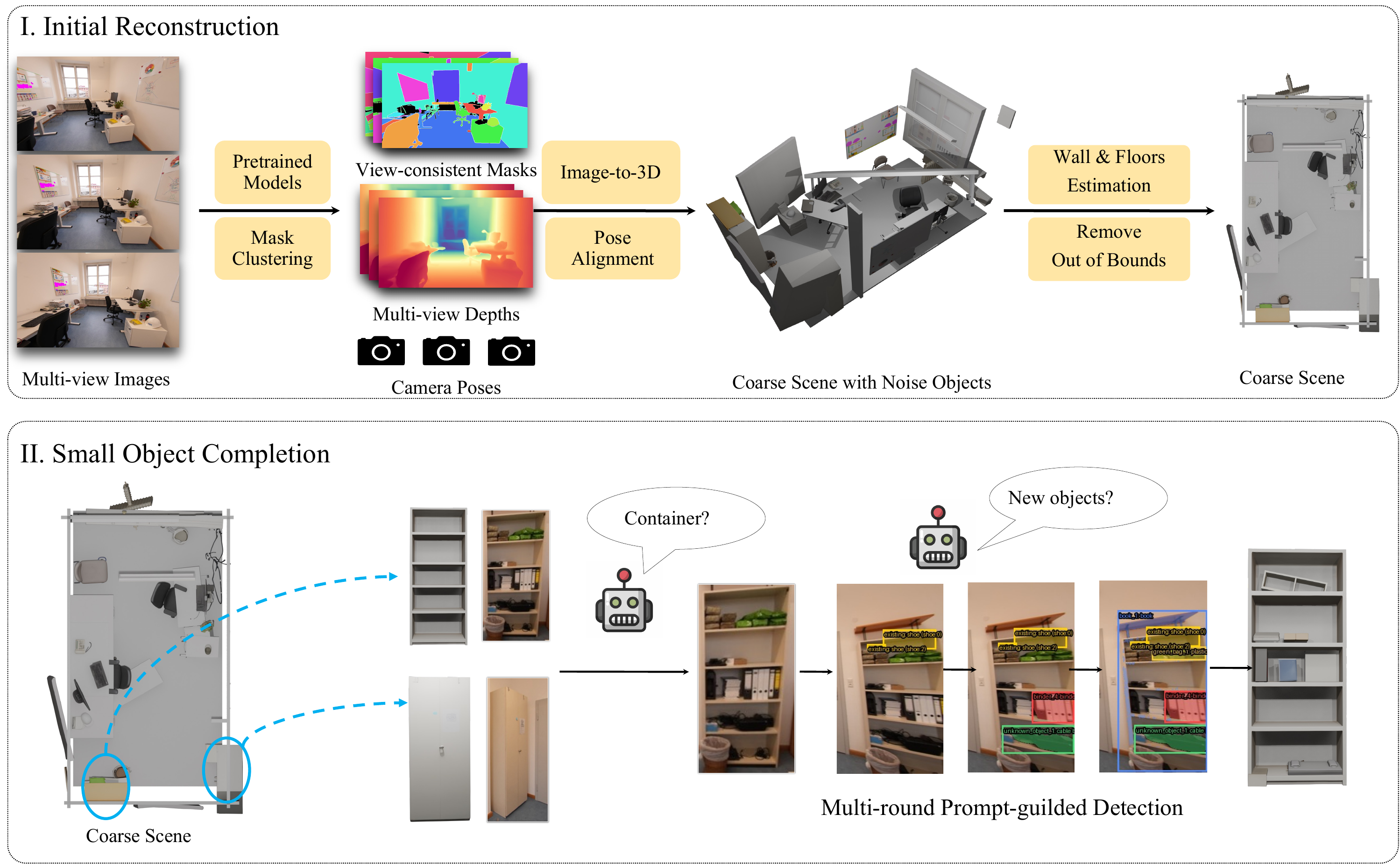}
    \caption{\textbf{Initial reconstruction with geometric and generative references.}
    VGGT-Omega supplies cameras, depth, and point maps, while CropFormer masks are clustered into 3D tracks. SAM3D generates and aligns a reference mesh from a selected view of each track; the agent then rebuilds the object with editable Blender primitives.}
    \label{fig:overview_3_1}
\end{figure}

Fig.~\ref{fig:overview_3_1} expands the reconstruction stage of Sec.~\ref{sec:method:enhance}.

\subsection{Geometry Reference Details}

\label{sec:appx:gen_align}
Our reference generation and alignment follow \citet{dong2026ras}, with the details below.

\vspace{1mm}\noindent\textbf{Generation.}
For object track $o$, let $\mathcal{K}_o$ contain the views with valid mask $M_v^o$. Rather than choose the largest 2D mask, which can favor a close-up showing only part of the object, we select the view with the greatest lifted surface area. For each $v\in\mathcal{K}_o$, we lift valid mask pixels with the VGGT-Omega point map, form a local triangular surface, and compute
\begin{equation}
    A_v^o = \sum_{\tau\in\mathcal{F}(M_v^o)} \operatorname{Area}(\tau),
    \qquad
    v^\star = \arg\max_{v\in\mathcal{K}_o} A_v^o,
\end{equation}
where $\mathcal{F}(M_v^o)$ denotes valid triangles induced by the lifted mask. This criterion favors views exposing more 3D surface and is less sensitive to camera distance than raw mask area.

We crop the RGB image, mask, and point map around $M_{v^\star}^o$ and condition SAM3D~\citep{sam3dobj}. The selected camera maps the generated canonical mesh back to scene coordinates, yielding an initial similarity transform $\Theta^{(0)}=\{s^{(0)},\mathbf{R}^{(0)},\mathbf{t}^{(0)}\}$. We use this single-view estimate only to initialize multi-view alignment.

\vspace{1mm}\noindent\textbf{Pose Alignment.}
Around the selected view, we form $\mathcal{V}_o=\{v^\star-r,\ldots,v^\star+r\}$ using only frames with a valid tracked mask. At iteration $i$, the mesh under $\Theta^{(i-1)}$ is rendered into each $v\in\mathcal{V}_o$, producing RGB $\widehat{I}_v^{(i)}$, depth $\widehat{D}_v^{(i)}$, and mask $\widehat{M}_v^{(i)}$. MASt3R~\citep{mast3r_eccv24} provides dense correspondences between the masked observation and rendering,
\begin{equation}
    \mathcal{C}_v^{(i)}
    = \{(\mathbf{p}_j,\mathbf{q}_j)\}_j
    = \Phi\!\left(I_v\odot M_v^o,\widehat{I}_v^{(i)}\right),
\end{equation}
where $\mathbf{p}_j$ and $\mathbf{q}_j$ are matched pixels in the observed and rendered images.

Let $\Pi_v^{-1}(\mathbf{p},d)$ back-project pixel $\mathbf{p}$ with depth $d$ using the VGGT-Omega intrinsics and camera-to-world pose. Each 2D match yields a 3D pair,
\begin{equation}
    \mathbf{x}_j = \Pi_v^{-1}\!\left(\mathbf{p}_j,D_v(\mathbf{p}_j)\right),
    \qquad
    \mathbf{y}_j^{(i)} = \Pi_v^{-1}\!\left(\mathbf{q}_j,\widehat{D}_v^{(i)}(\mathbf{q}_j)\right).
\end{equation}
Aggregating pairs over $\mathcal{V}_o$, we solve the incremental scale, rotation, and translation with Umeyama alignment~\citep{umeyama1991least}:
\begin{equation}
    \Delta\Theta^{(i)}
    = \arg\min_{\Delta s,\Delta\mathbf{R},\Delta\mathbf{t}}
    \sum_{(\mathbf{x},\mathbf{y})\in\mathcal{P}^{(i)}}
    \left\|\mathbf{x}-\left(\Delta s\,\Delta\mathbf{R}\mathbf{y}+\Delta\mathbf{t}\right)\right\|_2^2,
    \qquad
    \Theta^{(i)}=\Delta\Theta^{(i)}\circ\Theta^{(i-1)}.
\end{equation}
We repeat render--match--align for $K$ iterations and retain the transform with the highest mean rendered-to-tracked mask IoU:
\begin{equation}
    i^\star = \arg\max_{i\in\{1,\ldots,K\}}
    \frac{1}{|\mathcal{V}_o|}\sum_{v\in\mathcal{V}_o}
    \operatorname{IoU}\!\left(\widehat{M}_v^{(i)},M_v^o\right),
    \qquad
    \Theta^\star=\Theta^{(i^\star)}.
\end{equation}
\subsection{Primitive Representation Compactness}
\label{sec:appx:primitive_cost}
We quantify the compaction of Sec.~\ref{sec:method:enhance} on one office chair. The SAM3D reference has $178{,}172$ vertices and $356{,}084$ triangles; the agent reconstruction uses $43$ primitives ($22$ cuboids, $18$ cylinders, and $3$ frustums), $2{,}192$ vertices, and $1{,}183$ polygonal faces. This is an $81\times$ vertex and $301\times$ face reduction, and the authoring file drops from $22.9$~MiB to $610$~KiB after removing the reference mesh.
For geometric agreement, $20{,}000$ area-weighted reference samples have one-sided fitted-surface distances of $0.59\%$ of object height on average and $1.53\%$ at the $95$th percentile; silhouette IoU is $81.1$--$93.7\%$ over five orthographic views. This single-object measurement is against the generated reference, not the real chair.

\subsection{Small Object Completion}
\label{sec:appx:container_completion}

Scene-level segmentation can omit small objects inside containers. In a completion pass, the agent identifies reconstructed containers, inspects the corresponding image regions, and obtains masks for visible contents. REST3D~\citep{ma2026rest3d} then generates each missing object with SAM3D, while the container provides context for placing it back into the scene. This is part of reference-scene construction (Sec.~\ref{sec:method:enhance}), not a separately evaluated component.

\subsection{Deformable Simulation Settings}
\label{sec:appx:deformable_settings}

These settings accompany Tab.~\ref{tab:material_parameters}. All scenes use gravity $\mathbf{g}=(0,0,-9.81)~\mathrm{m/s^2}$; the telephone cord, chair cushion, and beanbag use $dt=0.01~\mathrm{s}$. For scenes with many deformables, rest-shape settling processes bodies in order of decreasing volume until its runtime budget is reached.

\noindent\textbf{Trial scene construction.}
Because IPC requires penetration-free input, each robot trial uses only the geometry needed for the interaction: (1) an inserted fixed-base Franka with seven revolute arm joints and two prismatic finger joints, modeled with affine body dynamics; (2) the target deformable and attached rigid bodies, such as the telephone handset and base; and (3) the remaining scene merged into a fixed collision environment.

\noindent\textbf{Telephone cord.}
The cord uses QIPC's native discrete elastic rod with twist~\citep{bergou2008discreteelasticrods}, initialized from the reconstructed centerline and radius. The reconstructed coil is the stress-free natural shape; Bishop-frame directors initialize the rod, and twist evolves dynamically. Stretching, bending, and twisting moduli are all $40~\mathrm{MPa}$. Together with the density and $1.3~\mathrm{mm}$ radius in Tab.~\ref{tab:material_parameters}, this gives a cord mass of about $14~\mathrm{g}$. These values come from an existing simulator phone example as a visual-plausibility prior and are not calibrated to the real cord.

The reconstructed centerline contains $3{,}200$ nodes over $1.694~\mathrm{m}$ and produces $15{,}755$ overlapping non-adjacent segment-capsule pairs, with up to $2.48~\mathrm{mm}$ overlap. To satisfy IPC's penetration-free input requirement, we fit a $576$-segment polyline whose segments all exceed the rod diameter by at least $1~\mu\mathrm{m}$. Segment lengths are $2.60$--$3.49~\mathrm{mm}$ (mean $2.89~\mathrm{mm}$), total discrete length is $1.664~\mathrm{m}$, and the maximum deviation from the reconstructed curve is $0.422~\mathrm{mm}$.

At each endpoint, the outermost three edges form a strain-relief lead attached to the handset or base: position and tangent are fixed in the body frame, while axial twist remains free. The handset and base are dynamic affine bodies of $0.2$ and $0.55~\mathrm{kg}$. Contact uses friction $0.6$, activation distance $0.15~\mathrm{mm}$, resistance $10^{8}$, and global continuous collision detection. Each step permits up to $100$ Newton iterations (velocity tolerance $2\times10^{-3}~\mathrm{m/s}$), $4{,}000$ PCG iterations (tolerance rate $10^{-6}$), and $24$ line-search iterations. We add no artificial damping or velocity modification.

Each $14~\mathrm{s}$ rollout stores $1{,}401$ states. The phone assembly starts $10~\mathrm{mm}$ above the scene and settles for $8~\mathrm{s}$. The gripper approaches from $8.6$ to $9.6~\mathrm{s}$, closes by $10.3~\mathrm{s}$, lifts the handset through finger contact from $10.7$ to $12.8~\mathrm{s}$, and holds until $14~\mathrm{s}$; independent finger PD controllers are limited to $20~\mathrm{N}$. A run passes if all segment stretches remain in $[0.9,1.1]$; settling drift and maximum nodal speed over $7.5$--$8~\mathrm{s}$ stay below $1~\mathrm{mm}$ and $2~\mathrm{mm/s}$; the handset rises and remains at least $120~\mathrm{mm}$ above its start, with hold slip below $5~\mathrm{mm}$ and finger--surface gap below $1~\mathrm{mm}$; rigid-body affine strain stays below $10^{-3}$; and robot joint residuals stay below $0.1~\mathrm{mm}$.

Both repeated runs pass without a failed state. Settling drift and speed are $0.773~\mathrm{mm}$ and $1.6~\mathrm{mm/s}$; peak handset lifts are $197.8$ and $197.7~\mathrm{mm}$. During lift, the minimum non-adjacent cord gap is $22.0~\mu\mathrm{m}$ in both runs, the minimum cord--environment gaps are $0.126$ and $0.110~\mathrm{mm}$, and endpoint error remains below $1.4\times10^{-14}~\mathrm{m}$. During hold, maximum finger-relative slip is $1.79$ and $1.77~\mathrm{mm}$ and the maximum finger--surface gap is $0.143~\mathrm{mm}$. No step exceeds $26$ Newton iterations. Segment stretch remains within $[0.9985,1.0463]$ and $[0.9983,1.0465]$.

The maximum stretch of about $1.046$ appears in the first step and remains nearly constant thereafter; lifting does not increase it. Resampling leaves $127$ next-but-one capsule pairs within the contact activation distance, the closest at $3.5~\mu\mathrm{m}$. At the first step, the barrier separates these pairs and lengthens $88$ short segments from $2.60$--$2.71~\mathrm{mm}$ to $2.72$--$2.74~\mathrm{mm}$, just below $2r+\delta=2.75~\mathrm{mm}$. Across the two repeated plans, peak lift, final handset height, settling drift, and settling speed remain within the declared repeat tolerances of $3~\mathrm{mm}$, $3~\mathrm{mm}$, $0.2~\mathrm{mm}$, and $0.5~\mathrm{mm/s}$; the peak lifts differ by $0.085~\mathrm{mm}$. The cord trajectories themselves differ by $0.665~\mathrm{mm}$ RMS and up to $9.55~\mathrm{mm}$ locally. The settled shape is written back to Blender as the rest shape (Sec.~\ref{sec:method:deform}); the natural shape remains the reconstructed coil, so the stored rest shape is an equilibrium under gravity rather than a stress-free configuration.

\noindent\textbf{Chair cushion.}
The seat and back cushions use StVK--Hencky solids for large compression; no viscous foam model is used. Their underside and rear surfaces are bonded to a fixed frame, with contact elsewhere. A $70~\mathrm{mm}$ rounded tool performs two tasks under one material model: a $1~\mathrm{s}$ gravity release and a $10~\mathrm{s}$ three-point press with $25~\mathrm{mm}$ commanded indentation. Both require principal stretches in $[0.1,3.0]$, minimum element Jacobian $0.04$, and at most $3~\mathrm{mm}$ displacement under gravity and after recovery; the press also requires $18$--$45~\mathrm{mm}$ peak indentation. Final indentations are $22.9$, $22.6$, and $22.9~\mathrm{mm}$.

\begin{figure}[t]
    \centering
    \includegraphics[width=0.65\textwidth]{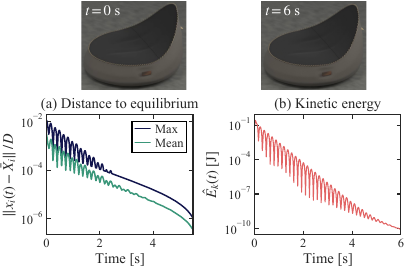}
    \caption{\textbf{Rest-shape settling.} Reconstructed beanbag at $t=0$ and $6$\,s. (a) Vertex distance to equilibrium, normalized by bounding-box diagonal $D=1.75$\,m; equilibrium is averaged over $t\in[5.5,6]$\,s. (b) Finite-difference kinetic energy. Maximum distance remains below $10^{-4}D$ after $2.3$\,s.}
    \label{fig:rest_shape_settling}
\end{figure}

\noindent\textbf{Beanbag settling.}
The beanbag uses stable Neo-Hookean solid finite elements with Tab.~\ref{tab:material_parameters}'s parameters. It is released under gravity without pinned vertices, added supports, or artificial damping; the surrounding scene is fixed and contact resistance is $10^{5}$. QIPC (v0.0.1.dev895) runs $600$ steps over $6~\mathrm{s}$ with $dt=0.01~\mathrm{s}$ and stores all $601$ states. Each step allows up to $100$ Newton iterations (velocity tolerance $0.005~\mathrm{m/s}$), $1{,}000$ linear-solver iterations, and $20$ line-search iterations. Fig.~\ref{fig:rest_shape_settling} shows convergence.
Relative to equilibrium, the initial geometry differs by at most $1.1\%$ of bounding-box diagonal $D$ and $0.29\%$ on average. Maximum vertex distance stays below $10^{-3}D$ after $1.15~\mathrm{s}$ and $10^{-4}D$ after $2.26~\mathrm{s}$. Finite-difference kinetic energy peaks at $0.26~\mathrm{J}$ and falls to $8.5\times10^{-11}~\mathrm{J}$ by $6~\mathrm{s}$. No tetrahedron inverts: minimum element Jacobian remains above $0.91$ and volume changes by less than $0.75\%$.

\noindent\textbf{Paper.}
Paper uses an isotropic St.\ Venant--Kirchhoff membrane with hinge bending: density $761.9~\mathrm{kg/m^3}$ (areal density $80~\mathrm{g/m^2}$), Young's modulus $2.25~\mathrm{GPa}$ for membrane and bending, Poisson's ratio $0.15$, thickness $0.105~\mathrm{mm}$, contact activation distance $0.5~\mathrm{mm}$, and friction $0.4$. QIPC's plastic variant uses yield curvature $25~\mathrm{m^{-1}}$ with no hardening, updating a yielded hinge's natural angle irreversibly. The elastic control disables bending plasticity and keeps initial natural angles; membrane plasticity is disabled in both.

The gripper executes a $140^\circ$ motion about a prescribed fold axis and pivot; this is the gripper motion, not a paper constraint. Both rollouts store $1{,}051$ states at $0.01~\mathrm{s}$ intervals, with opening from $7.2$ to $7.7~\mathrm{s}$. Fold angle is measured between area-weighted normals of two fixed material regions, with $0^\circ$ denoting parallel normals; the initial $5.35^\circ$ offset is not subtracted. Both rollouts initialize without detected intersections and pass strain and joint checks. The plastic run changes the natural angle of $372$ hinges by more than $0.1$~rad; the elastic run changes none. The elastic control was rerun after enabling plastic bending to isolate that model change, rather than using the original failed elastic trial. Fig.~\ref{fig:paper_fold_verification} and Tab.~\ref{tab:paper_fold} report the release behavior and fold angles.

\noindent\textbf{Plastic bag.}
The plastic bag in Fig.~\ref{fig:teaser} is a discrete thin shell with St.\ Venant--Kirchhoff in-plane elasticity and hinge bending: density $920~\mathrm{kg/m^3}$, Young's modulus $60~\mathrm{MPa}$ for membrane and bending, Poisson's ratio $0.45$, thickness $0.03~\mathrm{mm}$, and friction $0.45$. The bag deforms freely, without pinned or bound vertices, and the gripper holds it through contact alone. The trash bin is a separate fixed affine body with density $700~\mathrm{kg/m^3}$ and rigidity penalty $10^{9}$; the Franka uses rigidity penalty $10^{8}$. These values are simulation settings and are not calibrated to the real bag.

\noindent\textbf{Behavioral-test revisions.}
Tab.~\ref{tab:revision_log} summarizes revisions triggered while testing the three assets in Fig.~\ref{fig:paper_sim}.

\begin{table}[h]
\centering
\footnotesize
\setlength{\tabcolsep}{4pt}
\begin{tabularx}{\textwidth}{@{}l>{\raggedright\arraybackslash}Xl>{\raggedright\arraybackslash}Xc@{}}
\toprule
Asset & Observed failure & Attributed to & Revision & Material changed \\
\midrule
Telephone cord & No collision-free path in the first two plans & Motion & Re-planned trajectory & No \\
Cushion, indenter study & Element inversion and excess strain under large compression & Material model & Hencky logarithmic-strain elasticity & Yes \\
Cushion, press study & Element inversion and excess strain during the press & Motion & Revised press motion & No \\
Chair cushion & Back fragments fell under gravity; the solve did not converge & Geometry, boundary & Volume re-classified; full seat underside bonded & No \\
Chair cushion & Excess stretch and shallow indentation at the third point & Tool & $70$\,mm rounded tool & No \\
Chair cushion & Excess stretch at the third point & Contact location & Third point moved to the cushion front & No \\
Paper & Crease not kept after release & Material model & Plastic hinge bending ($\kappa_Y=25\,\mathrm{m^{-1}}$) & Yes \\
Paper & Plan check failed for the first two plans & Motion & Re-planned trajectory & No \\
\bottomrule
\end{tabularx}
\caption{\textbf{Revisions during agent-guided behavioral testing.} Rows are chronological within each asset. Earlier cushion studies supply the Hencky model and $E=30$\,kPa used in the final cushion; setup failures, repeats, interrupted runs, and rendering are omitted.}
\label{tab:revision_log}
\end{table}

\section{Evaluation Details}
\label{app:eval_details}

\subsection{Geometry and Rendering}
\label{app:eval_geometry}

\vspace{1mm}\noindent\textbf{Alignment.}
For the quantitative comparison, our method, HoloScene$^{*}$, and ReplicateAnyScene share VGGT-Omega cameras. We match estimated and dataset-provided reference camera poses by frame, initialize $\mathbf{x}_{\mathrm{ref}}=s\mathbf{R}\mathbf{x}_{\mathrm{pred}}+\mathbf{t}$ from camera centers with Umeyama alignment~\citep{umeyama1991least}, compute camera RMSE and rotation error, and refine the transform with scene-level ICP. GPT-6 Astra authors each scene in an independent coordinate frame with estimated scale and hand-placed cameras. We therefore register it by a global Z-up search over yaw in $5^\circ$ increments, with and without mirroring; initialize scale from horizontal room extents and translation from floor height and horizontal center; retain the candidate with the highest F-score at $10~\mathrm{cm}$; and run coarse-to-fine ICP with scale at $0.3$, $0.15$, $0.08$, and $0.05~\mathrm{m}$. Each transform is scene-level and applied to the full predicted mesh; no per-object alignment is used. Ground-truth meshes and reference camera poses are supplied in the HoloScene release; the ScanNet++ reference poses are estimated by SfM.

\vspace{1mm}\noindent\textbf{Geometry.}
We sample $500{,}000$ surface-area-weighted points from predicted mesh $P$ and ground-truth mesh $Q$ (seed $0$). Chamfer distance is the unsquared symmetric mean nearest-neighbor distance,
\begin{equation}
    \mathrm{CD}=\frac{1}{2}\Big(\frac{1}{|P|}\sum_{\mathbf{p}\in P}\min_{\mathbf{q}\in Q}\|\mathbf{p}-\mathbf{q}\|_2
    +\frac{1}{|Q|}\sum_{\mathbf{q}\in Q}\min_{\mathbf{p}\in P}\|\mathbf{q}-\mathbf{p}\|_2\Big),
\end{equation}
and is reported in centimeters. Precision (Prec) and recall (Rec) are the fractions of predicted and ground-truth points within $5~\mathrm{cm}$ of the other set, with $\mathrm{F1}=2\,\mathrm{Prec}\cdot\mathrm{Rec}/(\mathrm{Prec}+\mathrm{Rec})$ reported in percent.
Normal consistency (NC) is the mean absolute cosine similarity between the surface normal at each sampled point and that at its nearest neighbor in the other set, averaged over both directions and reported in percent. Evaluation includes walls, floors, ceilings, and other background surfaces: all render-enabled Blender surfaces are used for the prediction and the full mesh for ground truth, without foreground filtering or visibility culling.

\noindent\textbf{Rendering.}
We render every input frame: $811$, $371$, and $731$ views for the three Replica scenes, and $186$, $380$, and $528$ views for ScanNet++ scenes \texttt{67d702f2e8}, \texttt{7831862f02}, and \texttt{acd69a1746}. Frames are rendered at the input resolution ($512\times512$ for Replica and $876\times584$ for ScanNet++) with Cycles ($24$ samples, seed $0$), using the corresponding cameras in Blender coordinates. PSNR, SSIM, and LPIPS (AlexNet features) are computed on full unmasked images and averaged equally over frames. Because every method reconstructs from these same views, these metrics measure observation fidelity rather than held-out novel-view synthesis. Layout and Motion use $64$ uniformly sampled render/reference frame pairs (Appendix~\ref{app:eval_latent}).

\subsection{Rigid-Body Stability}
\label{app:eval_stability}

We use a MuJoCo rigid-body drop test as a physical-stability proxy.
For evaluation only, GPT-6 Astra assigns each reconstructed object to the same semantic categories for every method: walls, floors, and ceilings are \emph{background}; wall- or ceiling-mounted fixtures (e.g., paintings, lights, doors, windows, and attached blinds) are \emph{static}; and freestanding objects (e.g., furniture, books, plants, cushions, rugs, and bedding) are \emph{dynamic}. Background and static geometry remain fixed, while each dynamic object is tested individually under gravity.

\vspace{1mm}\noindent\textbf{Background.}
A standard convex decomposition would fill an enclosing room shell. We instead cluster triangles by face-normal direction (grouping angle ${<}\,20^{\circ}$), bisect each cluster until deviation from a fitted plane is below $3~\mathrm{cm}$, and extrude each patch into a $5~\mathrm{cm}$-thick convex slab, preserving the hollow interior.

\vspace{1mm}\noindent\textbf{Dynamic objects.}
Each dynamic object is decimated to at most $20{,}000$ faces and decomposed into at most $32$ CoACD~\citep{wei2022coacd} convex hulls (threshold $0.05$).

\vspace{1mm}\noindent\textbf{Per-object simulation.}
For each test, all other objects and static geometry are fixed. Simulation runs for $3~\mathrm{s}$ with $dt=0.002~\mathrm{s}$, gravity $-9.81~\mathrm{m/s^2}$, uniform density $300~\mathrm{kg/m^3}$, friction $1.0$, and an infinite ground plane at the estimated floor height. If a static background slab initially penetrates the test object by more than $2~\mathrm{cm}$, we remove that slab and restart, for at most three retries; this avoids artificial ejection from reconstruction overlaps such as a door embedded in its frame.

\vspace{1mm}\noindent\textbf{Fall criterion.}
An object counts as fallen if, after settling, either its local $z$-axis tilts by more than $30^{\circ}$ or its displacement exceeds $\max(5~\mathrm{cm},\;0.2d)$, where $d$ is its bounding-box diagonal.

\vspace{1mm}\noindent\textbf{Scores.}
Let $N$ be the number of dynamic objects, $N_g$ the number of \emph{ground objects} whose lowest point lies within $5~\mathrm{cm}$ of the floor, and $n_f$ and $n_f^{(g)}$ the numbers of fallen objects in the two sets.
\begin{equation}
    \text{Stable(All)} = 1 - \frac{n_f}{N},
    \qquad
    \text{Stable(Ground)} = 1 - \frac{n_f^{(g)}}{N_g}.
\end{equation}
Stable(All) also includes supported objects such as items on desks or shelves, so it depends on the reconstructed supports.

\subsection{Latent Similarity}
\label{app:eval_latent}

We use BVB's Latent Similarity (LS)~\citep{tang2026bvb}, comparing source and reconstructed videos in a frozen vision encoder's feature space.

\vspace{1mm}\noindent\textbf{Frame sampling.}
We uniformly sample $64$ one-to-one frame pairs from the input and rendered sequences, which share the same camera trajectory.

\vspace{1mm}\noindent\textbf{Encoding.}
Both clips pass through frozen V-JEPA~2.1~\citep{murlabadia2026vjepa21unlockingdense} ViT-G (\texttt{bf16}). Last-layer tokens are arranged on a $T\times H\times W$ grid, with each temporal step grouping two consecutive frames.

\vspace{1mm}\noindent\textbf{Motion.}
We average all tokens into one vector per clip; Motion is the cosine similarity between these vectors.

\vspace{1mm}\noindent\textbf{Layout.}
For Layout, tokens are averaged over time into a spatial feature map, and we average per-patch cosine similarity between the two maps. If tokens cannot form a rectangular spatial grid, Layout reduces to Motion.

\vspace{1mm}\noindent\textbf{LS.}
The combined score is $\text{LS}=(\text{Layout}+\text{Motion})/2$.

\vspace{1mm}\noindent\textbf{Remark.}
LS is affected by rendering style: our method uses fully lit Cycles renders, whereas ReplicateAnyScene uses unshaded vertex-colored meshes. We therefore treat LS as a complementary feature-space similarity measure, not as reconstruction quality alone or a direct test of downstream scene understanding.

\section{Reimplementations}
\label{app:reimpl}

We detail the baseline configurations and reimplemented components below. HoloScene$^{*}$ disables normal supervision and receives instance masks, cameras, and depth from our pipeline (Appendix~\ref{app:reimpl_holoscene}); ReplicateAnyScene and GPT-6 Astra start from RGB, with ReplicateAnyScene using its official VLM~+~SAM3 segmentation for the quantitative comparison (Appendix~\ref{app:reimpl_ras}).

\subsection{ReplicateAnyScene}
\label{app:reimpl_ras}

The official ReplicateAnyScene release omits the pose-alignment and VLM relation-reasoning modules described in the paper. We reimplement both and tune the reimplementation per scene.

\vspace{1mm}\noindent\textbf{Object segmentation.}
ReplicateAnyScene prompts a VLM for an object inventory from sampled frames, uses those names as SAM3~\citep{carion2025sam3segmentconcepts} text prompts, and propagates masks with SAM3-Video. The quantitative comparison in Tab.~\ref{tab:main_results} uses this default segmentation pipeline. The qualitative comparisons (Fig.~\ref{fig:scene_recon_vis} and Appendix~\ref{app:additional_vis}) instead show ReplicateAnyScene results obtained with manually annotated instance masks from the HoloScene release. For Fig.~\ref{fig:scene_recon_vis}, we also use ScanNet++'s SfM-estimated camera poses, matching the pose and mask sources of the official HoloScene result shown alongside it.

\begin{figure}[h]
    \centering
    \subfloat[Input images.\label{fig:hallway_input}]{%
        \includegraphics[width=\textwidth]{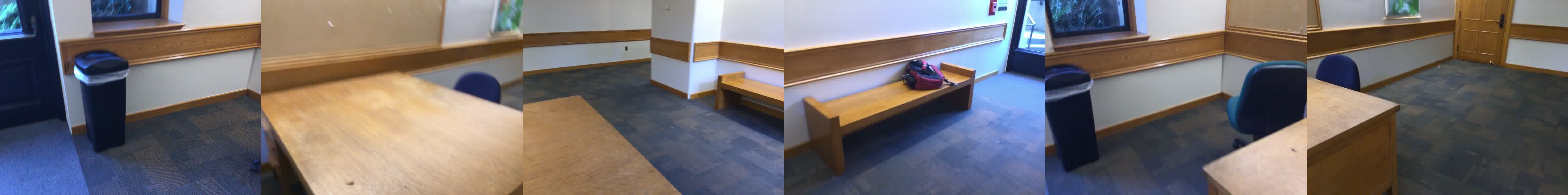}}\\[2pt]
    \subfloat[Reconstruction results (VLM~+~SAM3 masks).\label{fig:hallway_ras_recon}]{%
        \includegraphics[width=\textwidth]{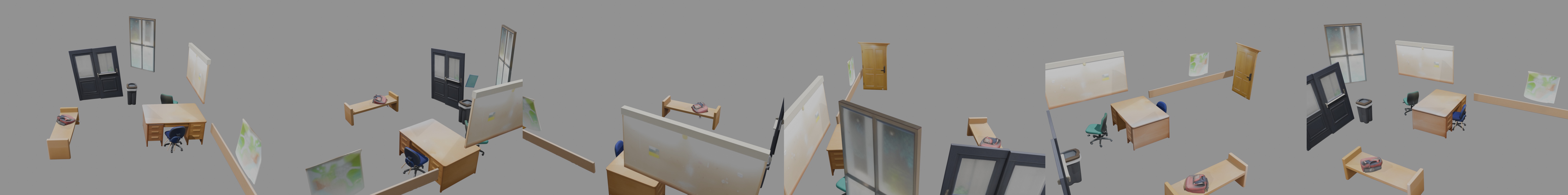}}
    \caption{\textbf{ReplicateAnyScene reimplementation on the official hallway example.} (a)~Six sampled input frames. (b)~Reconstruction with default VLM~+~SAM3 segmentation; the room shell is hidden to expose individual objects.}
    \label{fig:hallway_ras}
\end{figure}

\vspace{1mm}\noindent\textbf{Verification on the official example.}
As a sanity check, we run the full reimplementation on the hallway scene distributed with the official release. Fig.~\ref{fig:hallway_ras} shows the input frames and resulting object reconstruction.

\subsection{\texorpdfstring{HoloScene$^{*}$}{HoloScene*}}
\label{app:reimpl_holoscene}

HoloScene$^{*}$ in Tab.~\ref{tab:main_results} and Appendix~\ref{app:additional_vis} uses the official implementation with a modified configuration: normal supervision is disabled; instance masks are supplied by our mask-clustering pipeline (Sec.~\ref{sec:method:enhance}); and camera poses and per-frame depth are supplied by VGGT-Omega, as in CoDimRecon. These results characterize this modified configuration and should not be interpreted as the performance of HoloScene under its official default configuration.

The HoloScene result in Fig.~\ref{fig:scene_recon_vis} comes from the official release and is labeled without an asterisk. It uses ScanNet++'s SfM-estimated camera poses and the manually annotated instance masks provided in the HoloScene release.

\section{Prompts}
\label{app:prompts}
\label{app:prompt_gpt6_astra}

Listing~\ref{lst:gpt6_astra_prompt} gives the GPT-6 Astra baseline prompt from Sec.~\ref{sec:exp}. It is issued once per scene at reasoning effort \texttt{xhigh}. Only two strings vary: \texttt{<FRAME\_DIR>} points to the scene's RGB frames and \texttt{<OUTPUT\_DIR>} to its output directory. All other prompt text is identical across scenes and is provided verbatim.

\begin{lstlisting}[
  style=promptstyle,
  caption={Prompt used for the GPT-6 Astra scene-reconstruction baseline.},
  label={lst:gpt6_astra_prompt}
]
You are given a directory containing hundreds of RGB frames extracted from a video:

`<FRAME_DIR>`

You can only take the hundreds of RGB frames extracted from a video as input, do not take any other files as reference.

Your task is to autonomously reconstruct the **complete visible indoor scene** shown throughout the video as an explicit, object-decomposed, editable 3D scene in Blender.

This is a **full-scene reconstruction task**, not a single-object reconstruction task. Do not select only the most salient object, such as a desk, and stop after reconstructing it. You must recover the room structure and all visually significant, sufficiently observable objects across the video.

You are running on macOS. Blender is already installed. Use Blender through the command line and the Blender Python API. A typical Blender executable location is:

`/Applications/Blender.app/Contents/MacOS/Blender`

If necessary, locate the installed Blender executable yourself.

Use the following output directory:

`<OUTPUT_DIR>`

Create it if it does not exist.

Complete the entire task autonomously. Do not stop at planning, frame inspection, or the first modeling attempt.

## 1. Inspect the complete frame sequence

Inspect the frames in:

`<FRAME_DIR>`

The frames are temporally sampled from a video, so neighboring frames may be highly redundant. Efficiently inspect the complete sequence using scripts, image metadata, thumbnails, contact sheets, clustering, or other appropriate local tools.

Your inspection must cover the entire video timeline. Do not inspect only the beginning of the sequence or only frames containing one dominant object.

Determine:

* the overall room or hallway layout;
* which areas of the environment are visited by the camera;
* the major structural elements;
* all distinct, visually significant objects;
* repeated instances of the same object category;
* approximate spatial relationships among objects;
* which regions are well observed and which remain ambiguous;
* whether the video contains multiple connected areas or viewpoints of the same area.

Use image evidence rather than filenames alone.

## 2. Build a scene inventory before modeling

Before constructing the Blender scene, create:

`<OUTPUT_DIR>/scene_inventory.md`

The inventory must list every structural element and distinct physical object that should be reconstructed.

Include, where applicable:

* floor;
* walls;
* ceiling;
* doors and door frames;
* windows;
* openings;
* columns;
* baseboards and built-in architectural structures;
* desks and tables;
* chairs;
* cabinets and shelves;
* equipment and fixtures;
* lamps;
* screens, computers, keyboards, telephones, and similar devices;
* objects placed on furniture;
* cables or other visually meaningful thin structures;
* repeated object instances;
* any other significant objects visible in the scene.

For each item, record:

* a unique semantic name;
* object category;
* approximate scene location;
* the frames providing visual evidence;
* estimated geometric form;
* whether it is a unique object or a repeated instance;
* planned modeling method;
* reconstruction status;
* uncertainty or occlusion.

Do not collapse multiple distinct physical objects into one inventory entry merely because they share the same category.

The inventory must be created before detailed modeling and updated as reconstruction progresses.

## 3. Select primary reference views

Select at most 16 frames that collectively provide the most useful evidence for reconstructing the **entire scene**.

Save the exact absolute paths of these frames to:

`<OUTPUT_DIR>/selected_frames.txt`

Choose frames that jointly provide:

* coverage of different regions of the environment;
* diverse camera viewpoints;
* broad geometric coverage;
* complementary visibility of different surfaces;
* evidence for the room layout;
* evidence for object count and placement;
* clear and sharp observations;
* minimal occlusion;
* useful views of geometrically ambiguous regions.

Avoid near-duplicate neighboring frames.

Do not select all primary frames around only one object. The primary set must represent the complete visible environment as broadly as possible.

These frames should remain the primary reference set throughout reconstruction and refinement. Do not replace them merely to optimize the result after observing reconstruction errors.

If an object or geometric region remains ambiguous, occluded, or absent from the primary reference set, inspect additional frames as supplementary evidence.

Record every supplementary frame used, together with the reason it was needed, in:

`<OUTPUT_DIR>/supplementary_frames.txt`

There is no strict limit on supplementary frames, but avoid unnecessary redundant inspection.

## 4. Infer a coherent global scene layout

Before detailed object modeling, infer a consistent global coordinate system and approximate scene layout.

Establish:

* a consistent up direction;
* a ground plane;
* approximate global scale;
* room or hallway boundaries;
* major architectural dimensions;
* camera-traversed regions;
* approximate placement, orientation, and size of all major objects;
* support relationships, such as objects standing on the floor or resting on desks;
* repeated structures and repeated object spacing.

If exact metric scale cannot be recovered, choose a reasonable real-world scale based on familiar objects such as doors, desks, chairs, or computers, and document the assumption.

The final scene must be globally coherent. Do not model each reference frame as an unrelated local arrangement.

## 5. Reconstruct the complete scene in Blender

Reconstruct the environment as an explicit, editable, object-decomposed Blender scene.

The reconstruction is not complete merely because the most prominent desk or another dominant object has been modeled.

Reconstruct all visually significant and sufficiently observable elements identified in the scene inventory, including the architectural shell, major furniture, equipment, fixtures, and meaningful smaller objects.

### Object decomposition

Each distinct physical object should be represented as a separate, clearly named Blender object or logically organized collection whenever practical.

Use semantic names such as:

* `wall_left`;
* `door_01`;
* `desk_01`;
* `desk_02`;
* `chair_01`;
* `monitor_01`;
* `telephone_01`.

Do not merge the complete environment into a single undifferentiated mesh.

Organize the Blender scene into logical collections, for example:

* `Architecture`;
* `Furniture`;
* `Equipment`;
* `Small_Objects`;
* `Cables`;
* `Cameras`;
* `Lights`.

Repeated objects may share linked mesh data when appropriate, but each physical instance must have its own transform and identifiable object name.

### Geometry requirements

The final reconstruction should prioritize:

1. completeness of the visible scene;
2. correct global room layout and approximate scale;
3. correct object count;
4. correct object placement and orientation;
5. correct overall geometry and proportions;
6. correct spatial, support, and containment relationships;
7. correct major surface structure;
8. reasonable finer details supported by the images;
9. clean, editable, semantically organized Blender geometry.

Use real 3D geometry for important shapes. Do not rely on flat image planes, camera-facing billboards, or view-dependent geometry that only appears correct from one viewpoint.

The reconstructed scene must remain spatially plausible when viewed from novel viewpoints.

Avoid intersections between separate physical objects unless the images clearly support contact or overlap. Internal intersections between components belonging to the same assembled object are acceptable when reasonable.

### Materials

Assign reasonable materials and colors based on the visual evidence.

Exact photorealistic texture recovery is secondary to geometry, completeness, proportions, and layout. However, different semantic objects and visibly different surfaces should not all use one generic material.

Use procedural or simple image-based materials where appropriate. Keep materials editable and clearly named.

## 6. Use an iterative reconstruction workflow

Do not stop after generating the first scene.

Perform repeated reconstruction and verification cycles:

1. construct or modify the Blender scene;
2. render the current reconstruction from useful viewpoints;
3. inspect the renders;
4. compare them with the selected reference frames;
5. identify missing objects and structural elements;
6. identify incorrect object counts;
7. identify incorrect geometry, proportions, placement, or orientation;
8. correct the detected problems;
9. render and inspect again.

Every iteration must evaluate both:

* the accuracy of objects already reconstructed; and
* the completeness of the full scene.

Do not spend all refinement iterations polishing one desk or another individual object while other significant scene elements remain absent.

First obtain a complete coarse reconstruction of the entire scene. Then refine the geometry of individual objects according to visual importance and available evidence.

## 7. Render verification views

Create Blender cameras that provide useful views of the reconstructed scene.

Where camera poses cannot be recovered exactly, create approximate comparison views that show similar visible regions and viewing directions.

Produce:

* several scene overview renders;
* renders covering different regions of the environment;
* close-up renders of important or ambiguous objects;
* novel-view renders that help verify that the geometry is coherent in 3D.

Save verification renders under:

`<OUTPUT_DIR>/renders`

Do not rely only on Blender viewport screenshots. Render actual images through Blender.

After rendering, inspect the generated images yourself. Do not assume that a successfully executed render means the reconstruction is visually correct.

## 8. Perform a mandatory completeness audit

Before finishing, systematically compare the reconstructed Blender scene against:

* the full scene inventory;
* all selected primary frames;
* any supplementary frames used.

For every inventory item, assign one of the following final statuses:

1. `Reconstructed`;
2. `Partially reconstructed`;
3. `Not reconstructable from available evidence`;
4. `Intentionally omitted`.

For any item not fully reconstructed, document the specific reason. "Not a main object" is not a valid reason for omission.

Explicitly check for:

* missing structural elements;
* missing furniture;
* missing repeated instances;
* incorrect object counts;
* objects modeled in the wrong location;
* implausible floating objects;
* unintended intersections;
* inconsistent global scale;
* important objects represented only as vague placeholders;
* areas of the video that are not represented in the Blender scene.

A scene containing only the desk or another dominant object must be treated as incomplete unless the complete video genuinely contains no other reconstructable scene elements.

If significant omissions are detected, return to modeling and correct them before finishing.

## 9. Save the final deliverables

Save the final editable Blender scene to:

`<OUTPUT_DIR>/final_scene.blend`

Also save the Blender Python scripts used for reconstruction and rendering under:

`<OUTPUT_DIR>/scripts`

Save a final report to:

`<OUTPUT_DIR>/reconstruction_report.md`

The report must include:

* a summary of the reconstructed environment;
* the selected primary frames;
* supplementary frames consulted and why;
* the inferred global layout and scale assumptions;
* a complete list of reconstructed objects;
* the final status of every scene-inventory item;
* important geometric or material assumptions;
* remaining ambiguities;
* intentionally omitted elements and exact reasons;
* verification renders produced;
* the refinement iterations performed;
* the absolute path to the final `.blend` file.

## 10. Completion criteria

Do not declare the task complete until all of the following are true:

* the full video frame sequence has been inspected;
* the scene inventory has been created;
* the primary reference frames have been selected;
* the global scene layout has been established;
* the architectural structure has been reconstructed;
* all significant and sufficiently observable objects have been reconstructed;
* repeated objects are represented with the correct approximate instance count;
* the scene has been rendered from multiple useful viewpoints;
* the rendered results have been visually inspected;
* at least one substantive refinement pass has been completed after the initial full-scene model;
* a final completeness audit has been performed;
* the scene inventory and reconstruction report have been updated;
* the final editable Blender file has been saved successfully.

If some geometry remains ambiguous after consulting the available frames, make the most reasonable inference, model a plausible approximation, and document the uncertainty.

Work autonomously from start to finish. Continue inspecting, inventorying, modeling, rendering, comparing, and refining until you have produced the strongest complete-scene reconstruction possible under these constraints.
\end{lstlisting}

\section{Supplementary Demos}
\label{app:supp_demo}

The project page, \url{https://shuzhaoxie.github.io/CoDimRecon}, shows the reconstructed scenes and the deformable-object demos of Sec.~\ref{sec:exp}.

\section{Additional Visualizations}
\label{app:additional_vis}

\begin{figure}[!htbp]
    \centering
    \includegraphics[width=\textwidth]{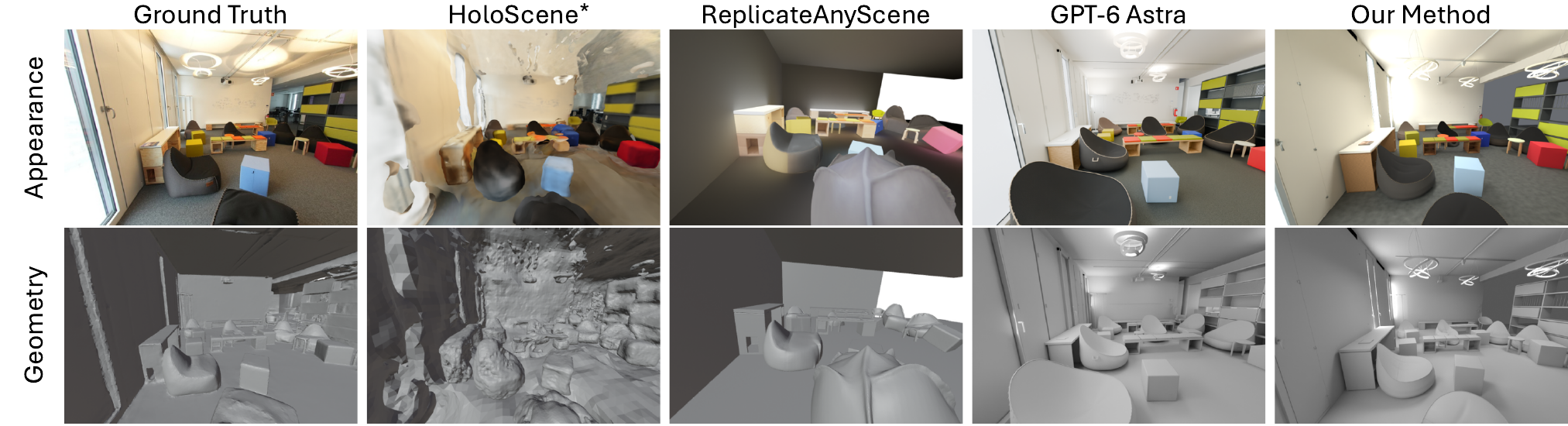}
    \caption{Qualitative comparison on ScanNet++ scene \texttt{7831862f02}. HoloScene$^{*}$ disables normal supervision and uses our estimated inputs (Appendix~\ref{app:reimpl_holoscene}).}
    \label{fig:add_vis_783}
\end{figure}

\begin{figure}[!htbp]
    \centering
    \includegraphics[width=\textwidth]{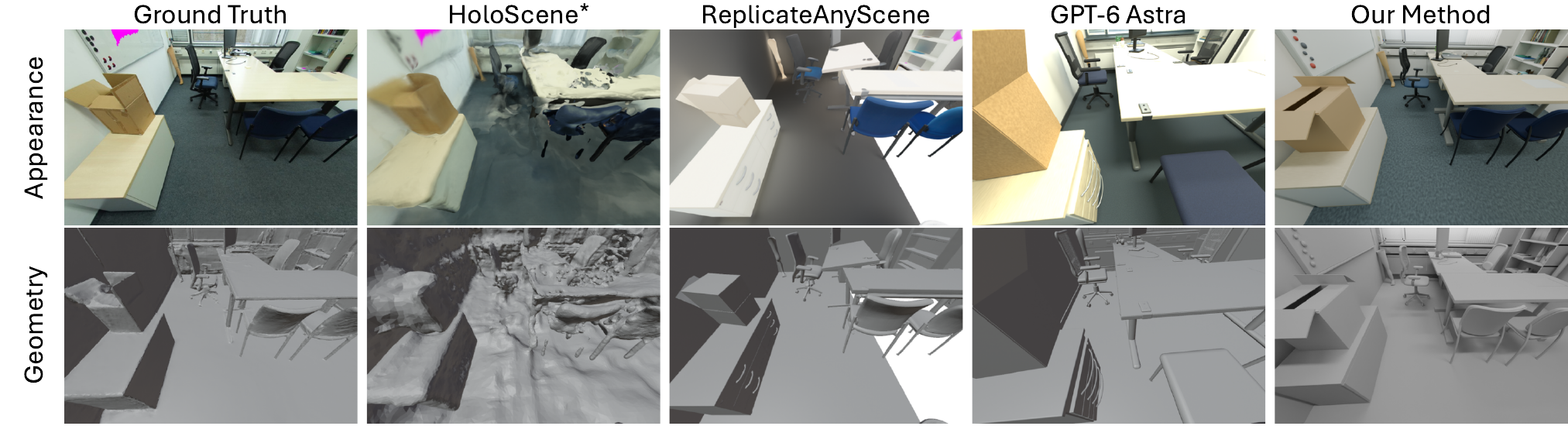}
    \caption{Qualitative comparison on ScanNet++ scene \texttt{acd69a1746}. HoloScene$^{*}$ disables normal supervision and uses our estimated inputs (Appendix~\ref{app:reimpl_holoscene}).}
    \label{fig:add_vis_acd}
\end{figure}

\begin{figure}[!htbp]
    \centering
    \includegraphics[width=\textwidth]{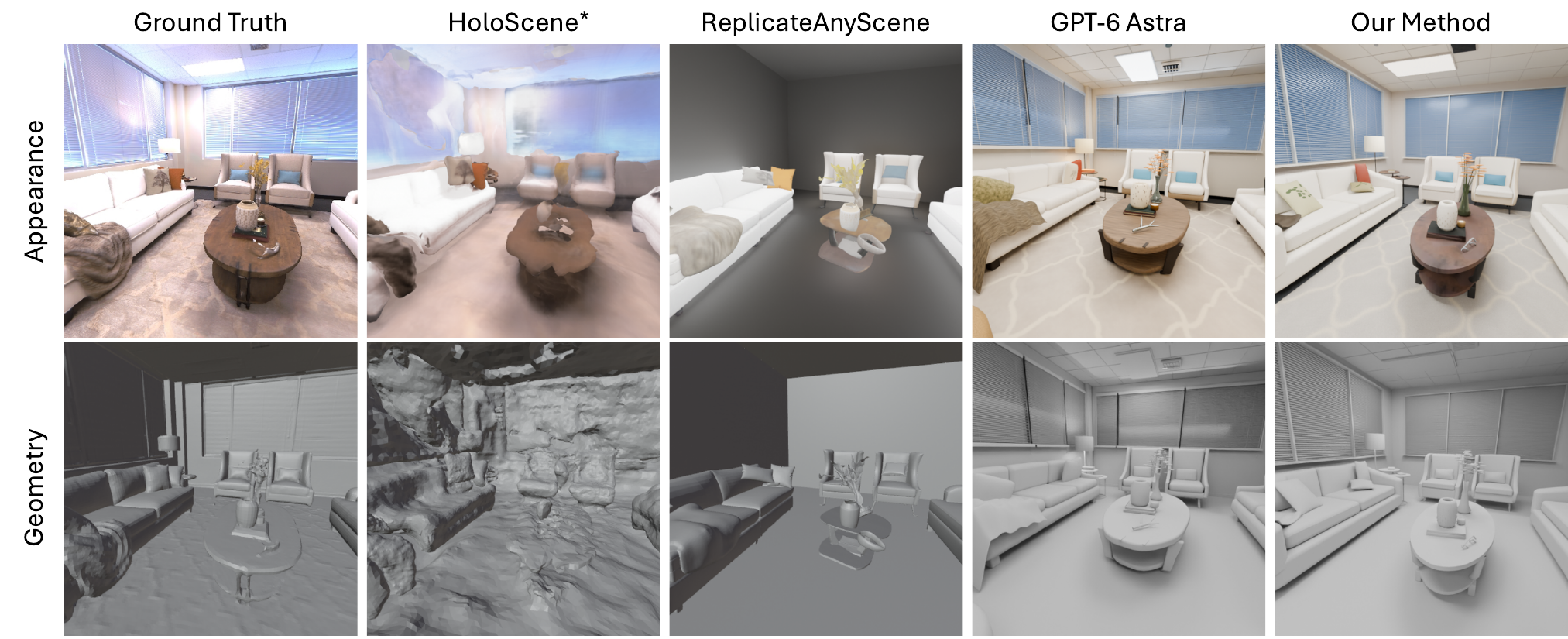}
    \caption{Qualitative comparison on Replica \texttt{room\_0}. HoloScene$^{*}$ disables normal supervision and uses our estimated inputs (Appendix~\ref{app:reimpl_holoscene}).}
    \label{fig:add_vis_room0}
\end{figure}

\begin{figure}[!htbp]
    \centering
    \includegraphics[width=\textwidth]{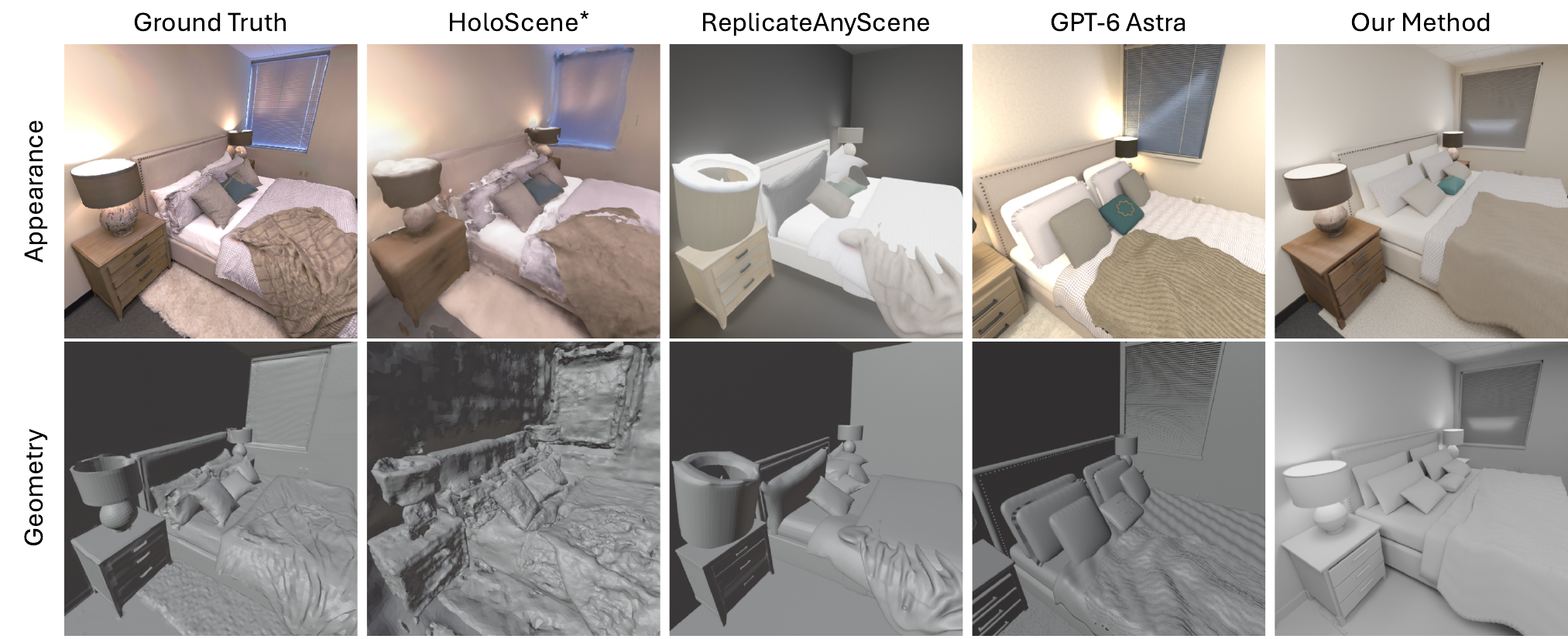}
    \caption{Qualitative comparison on Replica \texttt{room\_1}. HoloScene$^{*}$ disables normal supervision and uses our estimated inputs (Appendix~\ref{app:reimpl_holoscene}).}
    \label{fig:add_vis_room1}
\end{figure}

\begin{figure}[!htbp]
    \centering
    \includegraphics[width=\textwidth]{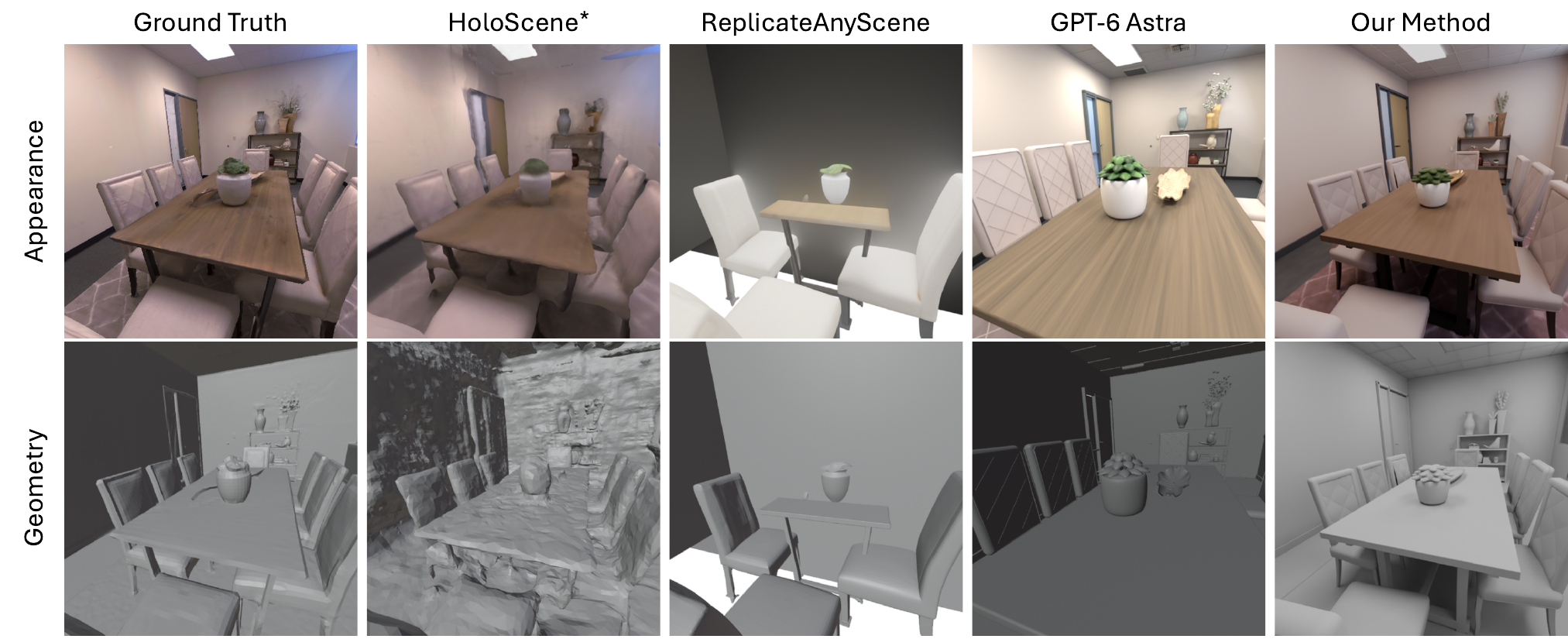}
    \caption{Qualitative comparison on Replica \texttt{room\_2}. HoloScene$^{*}$ disables normal supervision and uses our estimated inputs (Appendix~\ref{app:reimpl_holoscene}).}
    \label{fig:add_vis_room2}
\end{figure}

Figs.~\ref{fig:add_vis_783}--\ref{fig:add_vis_room2} provide additional qualitative comparisons
on the remaining ScanNet++ and Replica scenes, showing rendered
appearance (top) and geometry (bottom) for each method from the same input view. These figures show HoloScene$^{*}$ with normal supervision disabled (Appendix~\ref{app:reimpl_holoscene}). HoloScene$^{*}$ and our method use the same VGGT-Omega camera poses and depth and the same mask-clustering instance masks (Sec.~\ref{sec:method:enhance}).

\end{document}